\documentclass[preprint]{elsarticle}

\usepackage{graphicx}

\usepackage{hyperref}

\usepackage{amsmath}
\usepackage{amssymb}
\usepackage[linesnumbered,lined,boxed,commentsnumbered,ruled,norelsize]{algorithm2e}
\usepackage{multirow}
\usepackage{xcolor}
\usepackage{adjustbox}
\usepackage{fancyhdr}

\journal{Information Sciences}

\begin{document}
	
\begin{frontmatter}
\title{Pruning Deep Convolutional Neural Networks Architectures with Evolution Strategy}

\author[francisco]{Francisco E. Fernandes Jr.\fnref{fn1}}

\author[yen]{Gary G. Yen\corref{cor1}}
\ead{gyen@okstate.edu}

\address[francisco]{Institute of Mathematical Sciences and Computing, University of São Paulo, São Carlos, SP, 13566-590, Brazil}

\address[yen]{School of Electrical and Computer Engineering, Oklahoma State University, Stillwater, OK, 74078, USA}
\cortext[cor1]{Corresponding author}

\fntext[fn1]{This work was completed when the author was with the Oklahoma State University.}

© 2020. This manuscript version is made available under the CC-BY-NC-ND 4.0 license \url{http://creativecommons.org/licenses/by-nc-nd/4.0/}s

This preprint was accepted at the Information Sciences Journal with DOI \url{https://doi.org/10.1016/j.ins.2020.11.009}.

\begin{abstract}
Currently, Deep Convolutional Neural Networks (DCNNs) are used to solve all kinds of problems in the field of machine learning and artificial intelligence due to their learning and adaptation capabilities. However, most successful DCNN models have a high computational complexity making them difficult to deploy on mobile or embedded platforms. This problem has prompted many researchers to develop algorithms and approaches to help reduce the computational complexity of such models. One of them is called filter pruning, where convolution filters are eliminated to reduce the number of parameters and, consequently, the computational complexity of the given model. In the present work, we propose a novel algorithm to perform filter pruning by using Multi-Objective Evolution Strategy (ES) algorithm, called DeepPruningES. Our approach avoids the need for using any knowledge during the pruning procedure and helps decision-makers by returning three pruned CNN models with different trade-offs between performance and computational complexity. We show that DeepPruningES can significantly reduce a model’s computational complexity by testing it on three DCNN architectures: Convolutional Neural Networks (CNNs), Residual Neural Networks (ResNets), and Densely Connected Neural Networks (DenseNets).
\end{abstract}


\end{frontmatter}

\section{Introduction}
\subsection{Overview of the Problem}
Deep Convolutional Neural Networks (DCNNs) are among the most popular machine learning models used for computer vision applications at the moment. Although the basic theory behind DCNNs was developed in the late 1980s \cite{lecun_backpropagation_1989}, only recently, they were able to surpass humans in such tasks \cite{he_delving_2015, ioffe_batch_2015}. Mainly due to recent advancements in big data analytics and general-purpose computing on graphics processing units (GPUs).

DCNNs are data and model-dependent. In order to achieve good classification results, DCNNs need to be designed with millions of nodes, also called neurons, in its structure, requiring a massive amount of computational power to use them. Training such large DCNNs with small amounts of data can lead to overfitting problems where the model learns the training data very well, but it is unable to generalize its knowledge to unseen data. For example, \textit{ImageNet}, one of the most popular databases used for large-scale image classification, is composed of 14 million images distributed in $1,000$ classes \cite{deng_imagenet:_2009, russakovsky_imagenet_2015}. A typical DCNN capable of classifying such large databases may contain a few hundred million parameters. For instance, VGG-16, one of the top performer models in the ImageNet database in 2014, has $138$ million parameters \cite{simonyan_very_2014} and computational complexity of $15.5 \times 10^9$ floating-point operations ($15.5$ GFLOPS) \cite{li_person_2017}.

The task of finding a good DCNN architecture for a given problem, called Neural Architecture Search (NAS), is typically achieved by trial and error, and it can take months to find a suitable model depending on the amount of data and computing power available. Hence, in recent years, researchers have been using Evolutionary Computation (EC) and Swarm Intelligence (SI) tools to devise algorithms capable of automatically finding DCNN architectures and with little to no prior knowledge of the problem at hand \cite{darwish_survey_2020}. For example, Genetic Algorithms (GAs) were used with success to search for DCNN architectures by Yanan \textit{et al.} \cite{sun_evolving_2019, sun_evolving_2019-1}, Martín \textit{et al.} \cite{martin_evodeep_2018}, and many others \cite{real_large-scale_2017, miikkulainen_evolving_2017}. Particle Swarm Optimization (PSO), a particular SI algorithm, and Evolution Strategy (ES) were also successfully applied to the same task \cite{jiang_efficient_2019, fernandes_jr._particle_2019, shinozaki_structure_2015}.

However, most NAS algorithms are designed to find DCNN architectures with the best classification accuracy in a given problem. Thus, in general, DCNNs used for image recognition and classification tasks still require powerful hardware such as GPU workstations and datacenters for training and inference. Consequently, these models are not suitable for consumer hardware, such as smartphones and portable appliances. Currently, a constant connection to the internet is required if a developer is interested in performing computer vision tasks in such devices, which may not always be available, or may not be reliable to perform such tasks seamlessly to users. Thus, the development of algorithms and techniques that would reduce the computational complexity of DCNN models is highly desirable.

\subsection{Motivation}
Machine Learning researchers are increasingly interested in finding novel ways of reducing the computational complexity of DCNNs while maintaining their performance. There are mainly three approaches to achieve this goal: (1) designing DCNNs architectures from the ground up taking into consideration the limitations of the target platform; (2) making use of specialized mobile hardware capable of running DCNNs efficiently; and (3) modifying existing DCNNs architectures by removing redundant parameters. Each one of these techniques is detailed in the following.

In the first approach, novel DCNN operations and architectures can be created to take advantage of the particularities of mobile and embedded platforms. For instance, CondenseNets \cite{huang_condensenet:_2018} and MobileNets \cite{howard_mobilenets:_2017} are two specialized DCNN architectures created to be used on mobile platforms, such as smartphones. Both architectures make use of specialized convolutional operations, i.e., group and depth-wise convolutions, which decreases their computational complexity without degrading their performance. Likewise, NAS can also be used to find DCNN architectures suitable for mobile and embedded devices. The Differentiable Architecture Search (DARTS) algorithm developed by Liu \textit{et al.} is one example of a NAS algorithm capable of generating compact DCNNs by using a continuous search space \cite{liu_darts_2019}. The NetAdapt algorithm developed by Yang \textit{et al.} is another classic example capable of generating suitable DCNNs architectures given a certain resource budget. NetAdapt can find specific DCNN architectures tailored to an individual device \cite{ferrari_netadapt:_2018}. Another example is the FastDeepIoT algorithm developed by Yao \textit{et al.}, where an initial highly accurate model is found, and then the algorithm compresses the model by taking into consideration the execution time in a specific mobile and embedded device \cite{yao_fastdeepiot:_2018}.

The second approach to deal with the computational requirements of DCNN models is to develop specialized mobile and embedded hardware. One of the most iconic examples is the \textit{Jetson} embedded platforms developed by Nvidia, which features a GPU capable of performing inference in real-time using complex DCNN models \cite{rivera_nvidia_2017, su_constructing_2015}. Another area of intense interest is the use of Field-Programmable Gate Array (FPGA) for DCNN model inference. Some researchers have adapted deep learning frameworks and libraries to work with FPGA platforms resulting in inference time comparable with the ones from applying GPUs \cite{danopoulos_acceleration_2018, bottleson_clcaffe:_2016}. However, the use of such platforms does not address the problem of how to reduce the computational complexity of DCNN models. In that sense, Ding \textit{et al.} developed a method to reduce the computational complexity of Convolutional Neural Network (CNN) operations by using block-circulant weight matrices for implementation in FPGAs \cite{ding_circnn:_2017}. In their work, the weight matrices of each layer are converted to block-circulant matrices, and inference operations are performed using Fast Fourier Transform multiplications resulting in reduced computational complexity for inference and memory use.

Pruning is considered a third approach to reduce the computational complexity of DCNN models. In this case, the hypothesis is that many DCNN models handcrafted by humans have too many unnecessary parameters, and the identification and elimination of such parameters can reduce the model complexity without degrading its performance. Ding \textit{et al.} developed a technique called Auto-balanced Filter Pruning, where entire convolutional filters\footnote{In the present work, we refer to \textit{convolutional filters} or \textit{filters} as the trainable weights of a convolutional layer, while \textit{feature maps} refer to the output tensor of a convolutional layer.} can be removed from multiple layers based on their importance by using $l2$ regularization \cite{ding_auto-balanced_2018}. However, it requires that the initial DCNN model be trained with strong regularization, where weak filters are regularized with positive factors, while strong ones are regularized with negative factors. This initial training phase makes it easier to identify the filters to be removed during the pruning process, but it also requires the retraining of the original DCNN model used for pruning, which can take a long time. Luo \textit{et al.} \cite{luo_thinet:_2017} also performed pruning of DCNN models by removing entire filters from multiple layers. Their method is called ThiNet, and it uses statistics information from the next layer instead of the information from the current layer being pruned. The method developed by Li \textit{et al.} eliminates entire filters by computing the sum of the absolute values of the filters in a given layer, and removing the ones with the smallest sum \cite{li_pruning_2017}. Pruning has also been achieved with the use of \textit{Evolutionary Algorithms} (EAs). The works developed by Wang \textit{et al.} \cite{wang_towards_2018} and Zhou \textit{et al.} \cite{zhou_knee-guided_2019} are two recent examples in which EAs are used to prune DCNN models with competitive results.

It is also worth mentioning other approaches used to deal with the high computational demands of DCNN models. For example, it is possible to perform weight quantization, reducing the number of bits needed to represents the network’s weights \cite{han_deep_2016}. An extreme form of weight quantization is used in binarized networks in which all weights are represented using a single bit \cite{courbariaux_binaryconnect:_2015, hubara_binarized_2016}. Another approach is to design DCNN models that compute the residual mapping of a function, also known as a shortcut or a skip connection, used by Deep Residual Networks (ResNets) \cite{he_deep_2016}. The residual mapping avoids the problem of the vanishing gradient when creating models with hundreds or thousands of layers. Residual mapping does not directly reduce the computational complexity of a DCNN, but it helps in achieving high performance by using convolutional layers with smaller filter sizes. Densely Connected Convolutional Networks (DenseNets) \cite{huang_densely_2017} also use shortcut connections from previous layers, but, instead of using the output of only a single layer as in a ResNet, it uses the output from all previous layers as the input to the current one. In other words, each convolution layer in a DenseNet is connected to all subsequent ones in a fully-connected fashion. Similar to ResNets, the use of multiple shortcut connections between layers does not decrease the computational complexity of the network, but it allows the use of smaller networks with higher performance when compared with DCNNs with no shortcut connections.

We believe that all the works in building DCNN models with the best performance possible should not be discarded when trying to find models with reduced computational complexity. In this paper, we leverage such works by proposing a novel pruning strategy to eliminate unnecessary parameters from existing top-performing models. Specifically, we propose the elimination of redundant filters from convolutional layers to reduce the model’s computational complexity. This approach is considered the most suitable one for most applications because it eliminates the need to have specialized hardware to run such models. It also gives us some insight into the structure of DCNNs in general and how most models could be created using fewer parameters. In this sense, the pruning of a DCNN model can be seen as a combinatorial optimization where we try to find the optimal combination of convolutional filters in different layers from a given model. It can also be formulated as a multi-objective optimization problem where we have two conflicting objectives: (1) reduce the number of parameters in a given model as much as possible while (2) keeping the model’s performance as high accuracy as possible. We are also interested in prune the models without relying on any prior knowledge about them, allowing us to prune many different DCNN architectures with a single algorithm. The use of \textit{Evolutionary Computation} (EC) algorithms is ideal for this case. Such algorithms are well suitable for combinatorial optimization with conflicting objectives. Specifically, the \textit{Evolution Strategy} (ES) \cite{beyer_evolution_2002} is one such algorithm where a population of candidate solutions is evolved by performing random perturbations in their parameters.

\subsection{Contributions}
The main objective of this paper is to propose the use of the \textit{Evolution Strategy} for pruning DCNN models, which we call \textit{DeepPruningES}. We chose to use ES because of its simplicity, but powerful ability to find reasonable solutions. We highlight that we are interested in finding reasonable solutions instead of optimal ones. Thus, when pruning DCNN models, we can only find solutions with some trade-off between computational complexity and performance. The best solutions are in the so-called Pareto frontier, where no solution is strictly better than the others. However, the task of finding the true Pareto frontier for a given model pruning task is not trivial. Hence, the following is a list of contributions of the proposed \textit{DeepPruningES} algorithm:

\begin{itemize}
	\item The task of pruning DCNN architectures is conceived as a two-objective optimization problem and solved using an Evolution Strategy (ES), where the training error and the computational complexity are chosen to be two competing objectives. Furthermore, the proposed algorithm performs a gradual pruning of a chosen DCNN model, where filters are removed from the chosen DCNN model over the generations.
	\item The proposed algorithm provides three solutions with different trade-offs at the end of the pruning procedure and avoids the need to find the true Pareto frontier. In this regard, a Decision-Maker (DM) can choose the best solutions that fit his/her needs. These solutions are the following: 
	\begin{itemize}
		\item The candidate solution with the smallest training error in the population called the \textit{Boundary Heavy} solution;
		\item The candidate solution with the smallest computational complexity called the \textit{Boundary Light} solution;
		\item The candidate solution with the best trade-off between training error and computational complexity called the \textit{Knee} solution.
	\end{itemize}
	\item A modified ES selection operator based on the \textit{Minimum Manhattan Distance} (MMD) approach \cite{chiu_minimum_2016} is proposed to choose these three solutions from a population of candidate solutions. With this approach, the proposed algorithm does not require the use of any trade-off parameter.
\end{itemize}

The proposed algorithm is also distinct from others' pruning algorithms with evolutionary approaches, as explained in the following:
\begin{itemize}
	\item Wang \textit{et al.} \cite{wang_towards_2018} developed an evolutionary algorithm that, at first glance, looks similar to the proposed one. However, the algorithm in \cite{wang_towards_2018} does not incorporate a Multi-Criteria Decision Making (MCDM) framework, and can only find a single solution with a very small computational complexity regardless of classification accuracy, while the proposed algorithm can find three solutions with different trade-offs between computational complexity and classification performance. In the proposed algorithm, the DM is not required to set up any trade-off parameter as in \cite{wang_towards_2018}. The proposed pruning algorithm is developed as a Multi-Objective Optimization (MOP) algorithm in which two conflicting objective functions are optimized at the same time, while the algorithm presented in \cite{wang_towards_2018} uses a single objective function. Moreover, the authors in \cite{wang_towards_2018} do not make it clear if their algorithm is capable of pruning all layers of a ResNet or just layers between the shortcut connections. On the contrary, the algorithm presented in this work can prune all layers of ResNet and DenseNets by using two binary strings to encode their filters. Thus, we can prune the entirety of a ResNet or DenseNet model, not just its intermediate layers.
	\item At first glance, another algorithm that looks similar to the proposed one is the Knee-Guided Evolutionary Algorithm (KGEA) for Compressing Deep Neural Networks developed by Zhou \textit{et al.} \cite{zhou_knee-guided_2019}. The first novelty of the proposed algorithm compared with KGEA is that KGEA can only prune simple CNN models, such as VGG networks \cite{simonyan_very_2014}, while the proposed \textit{DeepPruningES} can prune multiple DCNN topologies, such as VGGs, ResNets, and DenseNets. Another difference is that KGEA tries to maintain a diverse Pareto front by using a mechanism to ensure diversity, while in \textit{DeepPruningES}, we are interested in finding good candidate solutions as fast as possible. Thus, \textit{DeepPruningES} does not require a diverse Pareto front to be maintained during the pruning process. Furthermore, because we are interested in preserving only three candidate solutions, using an ES framework is better suited for the problem than a GA one as does in \cite{zhou_knee-guided_2019}, which also results in a population with a smaller number of individuals than others' evolutionary approaches.
\end{itemize}

The proposed algorithm can prune Convolutional Neural Networks (CNNs), Residual Neural Networks (ResNets), and Densely Connected Neural Networks (DenseNets), which, to the best of our knowledge, no other algorithm proposed in the literature is capable of pruning. In general, it can achieve up to a 75\% reduction in the number of Floating-Point Operations (FLOPs) with little degradation to the model's classification performance.

The remaining sections of the present work are organized as follows. Detailed background about Deep Neural Networks models, such as CNN, ResNet, and DenseNet, Convolutional Filter Pruning, Evolution Strategies, is given in Section II. The proposed algorithm is presented in Section III. The experimental design used to validate the algorithm is discussed in Section IV. The experimental results are analyzed in Section V, followed by the conclusions and future work in Section VI.

\section{Background}

\subsection{Deep Convolutional Neural Networks}
A Deep Neural Network (DNN) is a computational model inspired by the way animal’s brains work. Its primary computational node is known as an artificial neuron, which is composed of a set of weights and a transfer function. To compute the output of a neuron, first, a weight operation is performed with its inputs and weights. Then, the result is passed through a non-linear function, known as activation function, to produce an output which can be used as inputs to subsequent neurons in a given DNN. In this sense, multiples neurons stacked together constitute a layer, and multiple layers stacked together constitute the architecture of a DNN. A Deep Convolutional Neural Network (DCNN) is, then, defined here as a DNN containing one or more layers using convolutional operations as their weight operations. Mathematically, a DCNN is defined as in Eq. \ref{eq:ann-equation}, where $i$ represents the layer number, $\mathbf{X}$ is a tensor representing the inputs of the entire DCNN, $\mathbf{O}_i$ is a tensor representing the output of layer $i$, $f_i(.)$ represents the activation function used in layer $i$, and $\mathbf{Z}_i$ is the output of the weight operation, $g_i(.)$, between the output from the previous layer, $\mathbf{O}_{i-1}$, and the weights of the current layer, $\mathbf{W}_i$. The three most commonly used weight operations are defined in Eq. \ref{eq:convolution}, and they are convolution, $\circledast$, max or average pooling, and matrix multiplication, used in fully-connected layers.

\begin{equation}
\label{eq:ann-equation}
\begin{cases} 
\mathbf{O}_{i} = \mathbf{X} & i = 1 \\
\mathbf{O}_{i} = f_{i}(\mathbf{Z}_{i}) & i > 1\\
\mathbf{Z}_i = g_i(\mathbf{O}_{i-1}, \mathbf{W}_i) \\
\end{cases}
\end{equation}

\begin{equation}
\label{eq:convolution}
\begin{cases} 
\mathbf{Z}_i = \mathbf{W}_i \circledast \mathbf{O}_{i-1} & \textnormal{if $i$-th layer is convolution}\\
\mathbf{Z}_i = Pool(\mathbf{O}_{i-1}) & \textnormal{if $i$-th layer is pooling}\\
\mathbf{Z}_i = \mathbf{W}_i \cdot \mathbf{O}_{i-1} & \textnormal{if $i$-th layer is fully-connected}\\
\end{cases}
\end{equation}

The convolutional layer was developed by LeCun \textit{et al.} in 1989 \cite{lecun_backpropagation_1989}, and allows DCNNs to go deeper due to its parameter sharing nature which reduces the total number of parameters when compared with an equivalent network constructed using only fully-connected layers. The pooling layer further decreases the number of parameters in a DCNN by subsampling its inputs with a max or average operation. Deep Convolutional Neural Networks (DCNNs) are constructed basically by stacking multiples convolution, pooling and fully-connected layers.

\begin{figure}[!t]
	\centering
	\includegraphics[width=0.95\columnwidth]{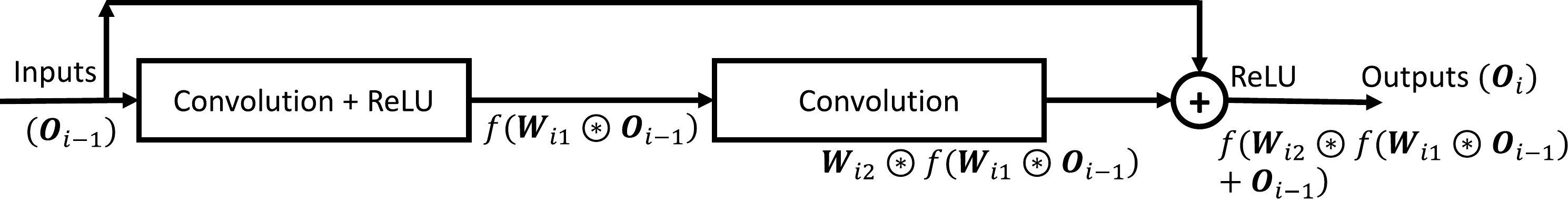}
	\caption{Example of residual block.}
	\label{fig:residual-block-example}
\end{figure}

\begin{figure}[!t]
	\centering
	\includegraphics[width=0.95\columnwidth]{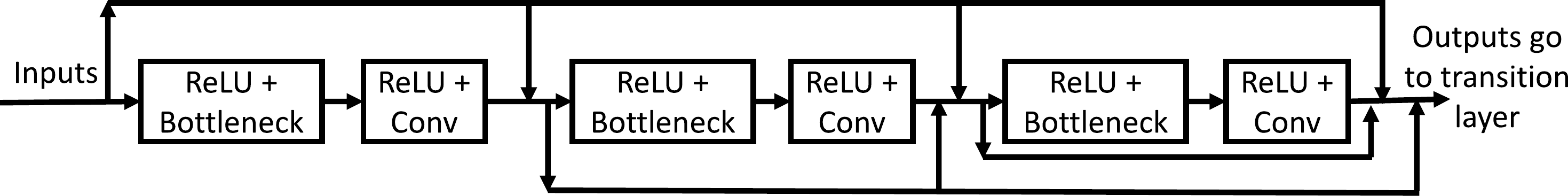}
	\caption{Example of a dense block with three bottleneck layers.}
	\label{fig:dense-block-example}
\end{figure}

In the case of Residual Neural Networks (ResNets) \cite{he_deep_2016}, their architecture is composed of stacked blocks, called residual blocks, which is a combination of two convolution layers with one shortcut connection. If we define the output of the previous residual block as $\mathbf{O}_{i-1}$, we can define the output of the current block, $\mathbf{O}_i$, as in Eq. \ref{eq:resnet}, where $f(.)$ is a rectified activation function, ReLU, i.e., $f(x)=max(0, x)$, $\mathbf{W}_{i1}$ are the weights of the first convolution layer, and $\mathbf{W}_{i2}$ are the weights of the second convolution layer. The visualization of such block can be seen in Fig. \ref{fig:residual-block-example}.

\begin{equation}
\label{eq:resnet}
\mathbf{O}_{i} = f(\mathbf{W}_{i2} \circledast f(\mathbf{W}_{i1} \circledast \mathbf{O}_{i-1}) + \mathbf{O}_{i-1})
\end{equation}

Densely Connected Neural Networks (DenseNets) \cite{huang_densely_2017} are also composed of blocks, called \textit{Dense Blocks}, which consist of multiples convolution layers. Their main difference compared with ResNets is that the outputs of each convolution layer in a given \textit{Dense Block} is used as input for all subsequent convolution layers from the same \textit{Dense Block}. The number of output feature maps from any given convolution layer is called \textit{growth rate}, $k$. Because the outputs from previous layers are concatenate and used as the input of the current layer, the number of input feature maps in layer $i$ is equal to $k_0 + k \times (i-1)$, where $k_0$ is the number of output feature maps from the first layer. Furthermore, in order to improve computational efficiency, each layer in a \textit{Dense Block} is composed of a bottleneck layer followed by a convolution layer. A bottleneck layer is simply a convolution layer with $1 \times 1$ convolution filters which receives all the inputs from previous layers. A transition layer is used between each \textit{Dense Block} with the purpose of reducing the number of inputs to the next block. The transition layer is normally built by using a pooling layer followed by $1 \times 1$ convolution layer. Fig. \ref{fig:dense-block-example} illustrates the connections between layers in a \textit{Dense Block} with three bottleneck layers.

\subsection{Convolutional Filter Pruning}
The weights of a convolution layer, also known as filters or kernels, can be represented as a four dimensional tensor $\mathbf{K}$ with dimensions equal to $(O \times D \times W \times H)$, where $O$ is the total number of kernels with size $(D \times W \times H)$, $D$ is the depth of each kernel, $W$ is the width of each kernel, and $H$ is the height of each kernel. Optionally, each one of the $O$ kernels can have a single parameter to represent its bias. If one would like to reduce the computational complexity of a single convolution layer, one approach would be to eliminate some of these $O$ kernels in their entirety. This approach is indeed one of the most common techniques used to reduce the computational complexity of DCNN models, and it is known as filter pruning. This process is illustrated in Figs. 3 and 4 where a filter is first chosen to be eliminated, indicated by the dashed lines in Fig. 3, and, then, the entire DCNN model is adjusted resulting in a pruned model, shown in Fig. 4. It is important to note that when filters are eliminated in layer $i$, the corresponding depths of all filters in the subsequent layer $i+1$ need to be removed to produce a valid architecture.

\begin{figure*}[!t]
	\centering
	\includegraphics[width=0.8\textwidth]{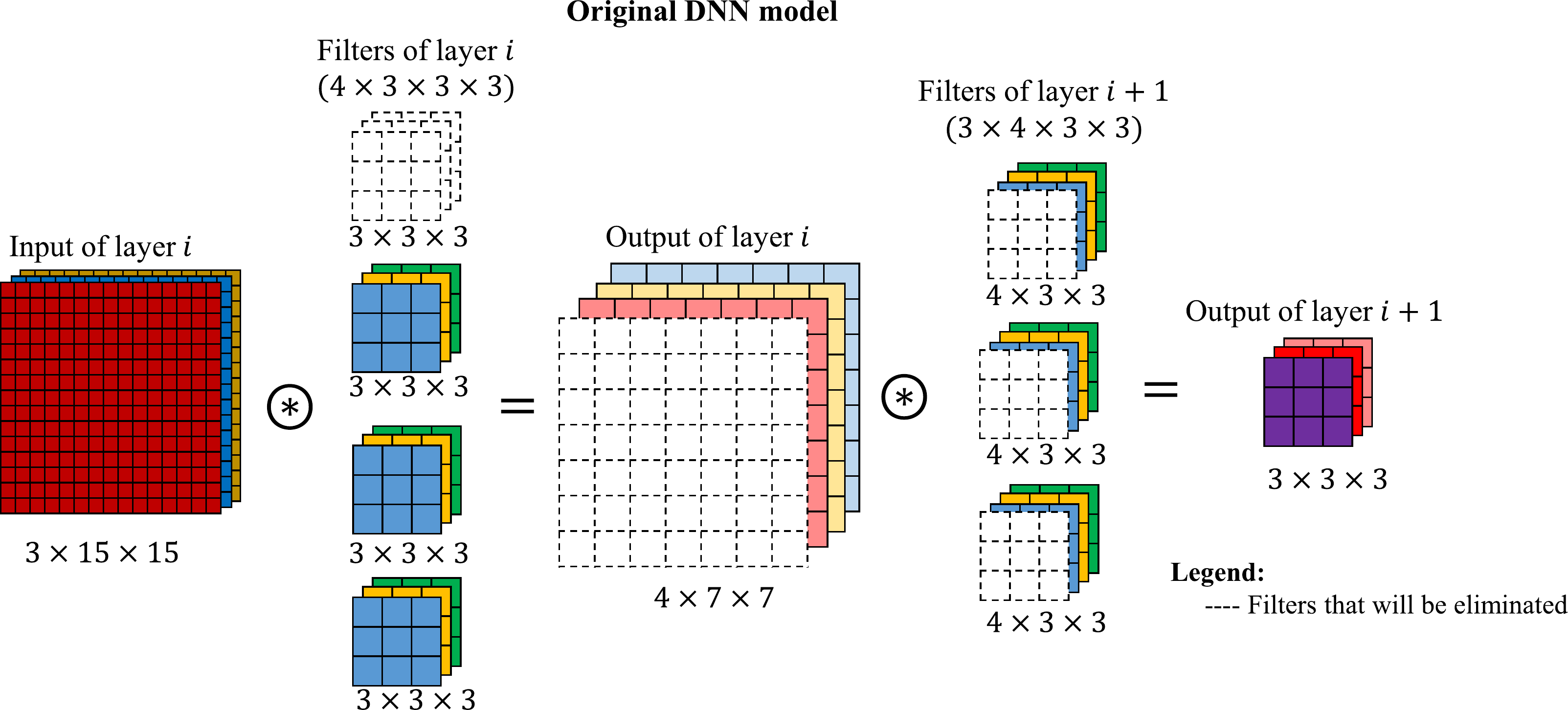}
	\caption{Example of convolutional filter pruning where the first filter of the current convolution layer is selected to be eliminated.}
	\label{fig:filter-pruning1}
\end{figure*}

\begin{figure*}[!t]
	\centering
	\includegraphics[width=0.74\textwidth]{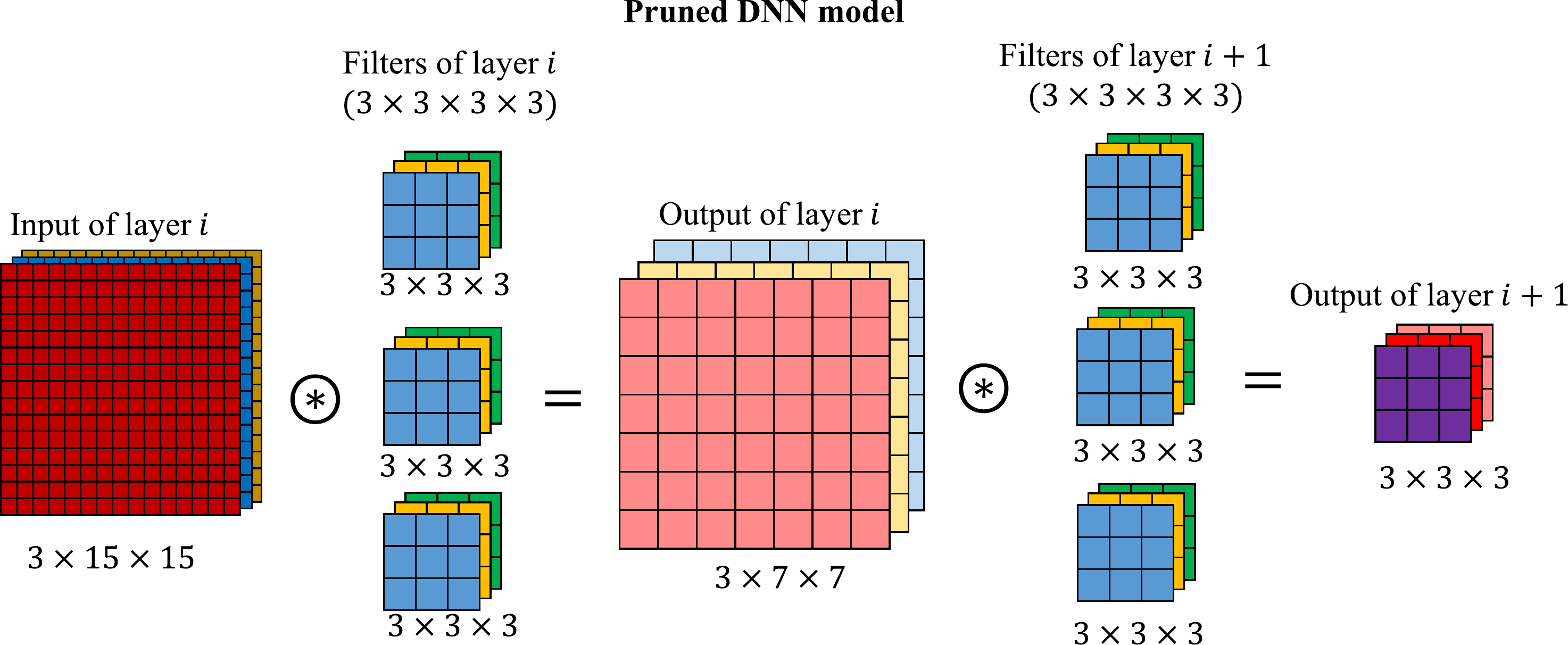}
	\caption{Pruned DCNN model after the first filter is removed.}
	\label{fig:filter-pruning2}
\end{figure*}

In the literature, there are multiple ways of choosing which filters to eliminate in each convolution layer, a procedure known as \textit{filter selection}. In general, some statistics from the current layer is needed to determine which filters to eliminate. The following are some of the most used decision criteria found in the literature \cite{mittal_studying_2019}:

\begin{itemize}
	\item \textit{Mean activation}: It was used by Molchanov \textit{et al.} in \cite{molchanov_pruning_2016} where the mean activation of each filter from a given layer is computed, and only the filters with the highest mean activation are kept.
	\item $l1$-norm: It was used by Li \textit{et al.} in \cite{li_pruning_2017} where the importance of a filter is determined by its $l1$-norm. Filters with smaller $l1$-norm are considered unimportant compared with filters with large $l1$-norm.
	\item \textit{Average Percentage of Zeros (APoZ)}: It was proposed by Hu \textit{et al.} in \cite{hu_network_2016}, and it eliminates the filters with a large number of zero activations during inference time.
	\item \textit{Entropy}: In this case, the entropy of a single filter is computed by feeding input images to the network and grouping the corresponding outputs in different bins. The key idea is that unimportant filters will have very similar outputs with different inputs. It was proposed by Luo and Wu in \cite{luo_entropy-based_2017}.
	\item \textit{Random}: It randomly select filters to be eliminated without using any prior knowledge of the DCNN architecture or filters. In \cite{mittal_studying_2019}, Mittal \textit{et al.} showed that eliminating filters at random produced comparable results with other selection criteria.
\end{itemize}

\subsection{Evolution Strategies}
The work done by Mittal \textit{et al.} in \cite{mittal_studying_2019} shows that filter pruning with randomized heuristics produces results as good as when using heuristics based on prior knowledge. Moreover, the use of randomized heuristics was capable of producing excellent results even when used for automatic generation of DCNN architectures \cite{liu_hierarchical_2017}. Thus, the use of an algorithm that relies on random changes of candidate solutions is ideal for performing filter pruning.

The evolutionary computation (EC) family of algorithms possess such characteristic. Specifically, Evolution Strategies (ESs) \cite{beyer_evolution_2002} relies more heavily on the use of random changes than any other EC algorithms. ESs are meta-heuristic algorithms used to solve optimization problems. Initially, ESs were created to solve design problems automatically, but its creators quickly realized that they could be used to solve all sorts of optimization problems, such as binary optimization and combinatorial optimization. Similar to other EC algorithms, ESs are population-based algorithms where an individual in the population is a possible solution.

In general, ESs work in a loop, known as generations, where these two basic steps are performed \cite{beyer_evolution_2002}:
\begin{itemize}
	\item In every generation, modify the individuals’ parameters by using recombination and/or mutation.
	\item Keep the best individuals for the next generation and repeat the process.
\end{itemize}

This family of algorithms uses the following standard notations: $(\mu / \rho , \lambda)-ES$ or $(\mu / \rho + \lambda)-ES$, where $\mu$ represents the number of individuals that will be kept and used as parents to the next generation, $\lambda$ represents the number of individuals that will be generated at every new generation from the $\mu$ parents selected in the previous generation, and $\rho$ is the number of individuals used for recombination. The \textit{comma} (,) and \textit{plus} (+) operators define how the selection operator of the algorithm will work. In the \textit{comma} version, the algorithm will select $\mu$ individuals only from $\lambda$ offspring, and the old parents will be completely discarded even if they are better than the offspring. In the \textit{plus} version, the algorithm will select $\lambda$ individuals from all $\mu + \lambda$ individuals, i.e., parents and offspring are used together during the selection procedure. Thus, the \textit{plus} version is an elitist algorithm where the best solutions are kept in the population indefinitely. Recombination is an optional step in ESs and it works by randomly selecting $\rho$ individuals from $\lambda$ parents. In ES, recombination always produces one new individual from two or more parents, and each one of its parameters are selected from one of the parents at random. Let $\mathbf{P}_j$ be a vector representing the $N$ parameters of a parent individual $j$, then the parameters of a new offspring, $\mathbf{O}$, will be equal to
\begin{equation}
\label{eq:es-recombination}
{O}_{k} := {rand}_j (P_{jk}), \textnormal{ with } 1 \leq k \leq N \textnormal{ and } j = 1, ..., \rho. 
\end{equation}

In the present work, DCNN models are pruned with the help of ES and without the use of any prior knowledge. In our case, the selection pressure produced by the evolutionary process will be responsible for finding a good combination of filters to be kept in a pruned DCNN model.

\section{Proposed Algorithm}

The general framework of the proposed \textit{DeepPruningES} is shown in Algorithm \ref{alg:DeepPruningES}. It is comprised mainly of the standard ES \textit{plus} version adapted to perform DCNN pruning. It receives as input the size of the offspring population ($\lambda_{size}$), the maximum number of generations that the algorithm will run ($gen$), the original DCNN model ($model$) that will be pruned, the number of epochs to be used during an individual evaluation ($e_{eval}$), the learning rate used during an individual evaluation ($\alpha_{eval}$), and the number of epochs ($e_{fine}$) and learning rate ($\alpha_{fine}$) to be used during the fine-tuning of the best individuals found at the end. The final output of the proposed \textit{DeepPruningES} will be three DCNN models with different trade-offs called here as \textit{knee}, \textit{boundary heavy}, and \textit{boundary light} solutions.

There are four important pieces in Algorithm \ref{alg:DeepPruningES}: \textit{Population Initialization} (line 1), \textit{Knee and Boundary Selection} (line 4), \textit{Offspring Generation} (line 5), and \textit{Fine-Tuning} of the best solutions found (line 7). Each one of them will be explained in detail in the following subsections, as well as the genetic representation of an individual.

\begin{algorithm}[!t]
	\caption{Proposed \textit{DeepPruningES}}
	\label{alg:DeepPruningES}
	\SetKwInOut{Input}{Input}\SetKwInOut{Output}{Output}
	
	\Input{Offspring size ($\lambda_{size}$), maximum number of generations ($gen$), mutation probability ($p_m$), original DCNN model ($model$), number of epochs for individual evaluation ($e_{eval}$), learning rate for individual evaluation ($\alpha_{eval}$), number of epochs for fine-tuning ($e_{fine}$), learning rate for fine-tuning ($\alpha_{fine}$).}
	\Output{Three DCNN models: knee solution ($\mu.knee$), boundary heavy solution ($\mu.heavy$) and boundary light solution ($\mu.light$).}
	
	\BlankLine
	$\mu, \lambda \leftarrow $ \textit{Initialize Population}($\lambda_{size}$, $model$)\;
	
	\For{g = 1 to $gen$}
	{
		$\mathbf{P} \leftarrow \boldsymbol{\mu} + \boldsymbol{\lambda}$
		\BlankLine
		$\boldsymbol{\mu} \leftarrow $ \textit{Knee Boundary Selection}($\mathbf{P}$, $model$, $e_{eval}$, $\alpha_{eval}$)\;
		$\boldsymbol{\lambda} \leftarrow $ \textit{Offspring Generation}($\boldsymbol{\mu}$, $\lambda_{size}$, $p_m$, $model$)\;
	}
	\textit{Fine-tuning}($\boldsymbol{\mu}$, $model$,  $e_{fine}$, $\alpha_{fine}$)\;
	\textbf{return} $\mu.knee$, $\mu.heavy$, $\mu.light$\;
\end{algorithm}

\subsection{Genetic Representation of an Individual}

Representation is one of the most important aspects of any population-based algorithm. In the proposed \textit{DeepPruningES}, we use two binary strings to represent the filters of a DCNN model. It is important to note that our proposed algorithm only prunes convolution layers because they have a much higher computational complexity compared to fully-connected ones \cite{li_pruning_2017}. In this representation, each bit represents one single filter. For example, if we want to represent two convolution layers, one with 32 filters and the other with 64 filters, we would need a binary string with 96 bits, and a bit assigned to zero means that the corresponding filter would be eliminated.

\begin{figure}[!t]
	\centering
	\includegraphics[width=0.9\columnwidth]{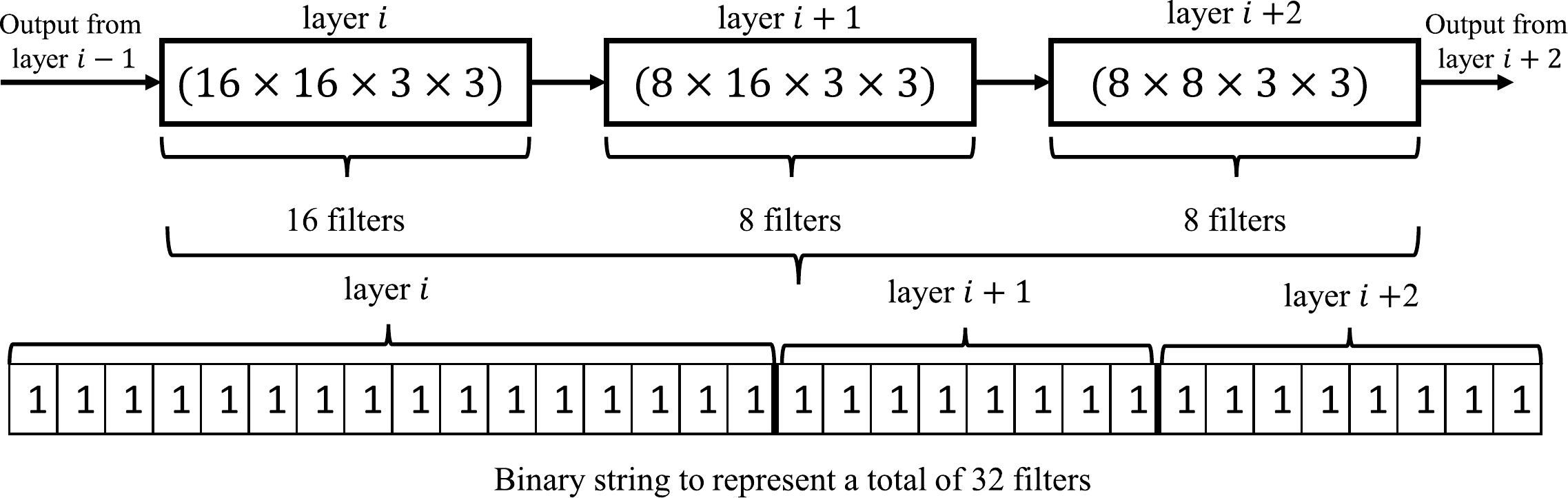}
	\caption{Binary representation of a CNN model. All layers are convolution layers.}
	\label{fig:CNN-representation1}
\end{figure}

\begin{figure}[!t]
	\centering
	\includegraphics[width=0.9\columnwidth]{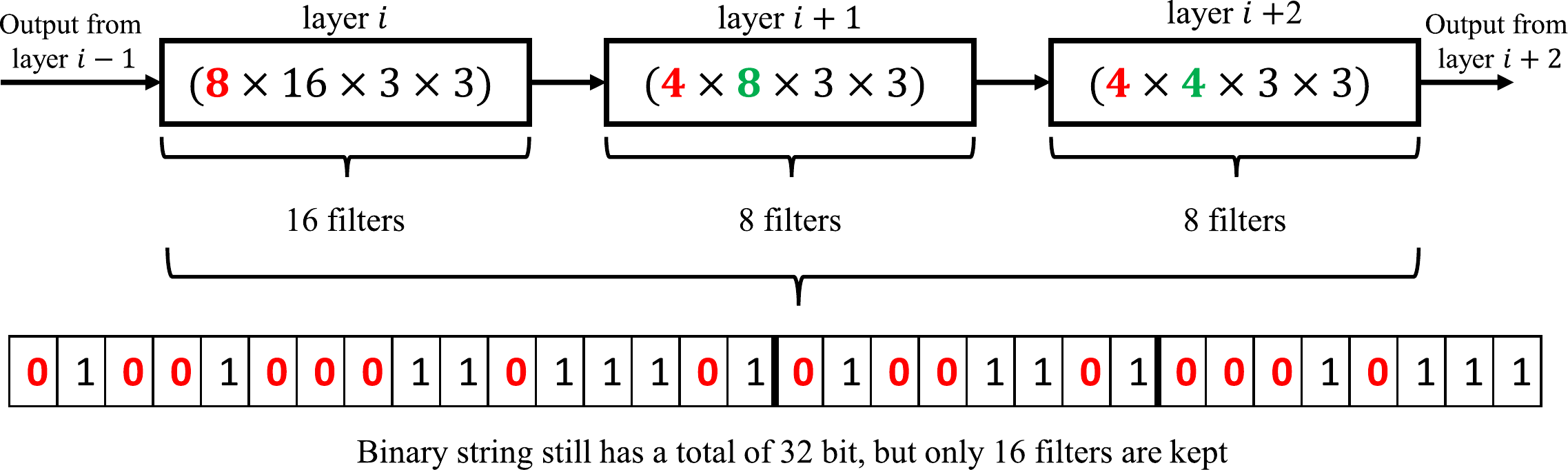}
	\caption{Binary representation of a CNN model after removing half of the filters from each layer.}
	\label{fig:CNN-representation2}
\end{figure}

\begin{figure}[!t]
	\centering
	\includegraphics[width=0.45\columnwidth]{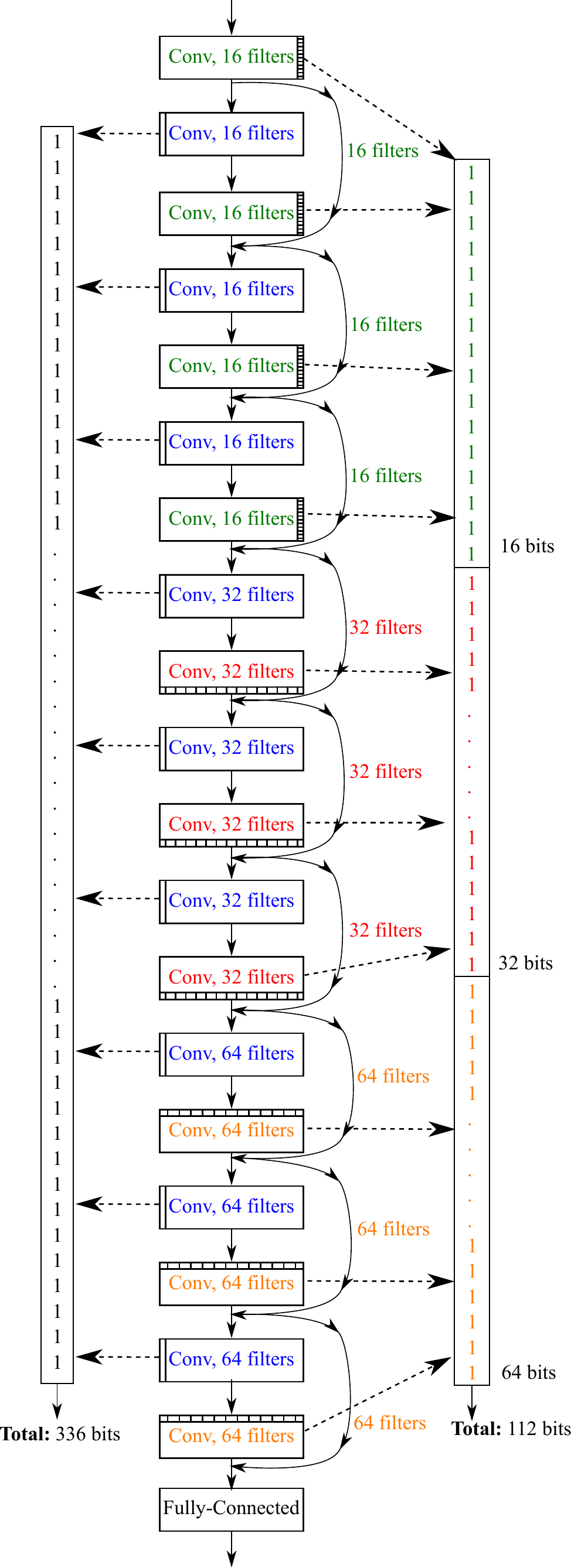}
	\caption{Binary representation of a ResNet with 20 layers.}
	\label{fig:resnet-representation}
\end{figure}

\begin{figure}[!t]
	\centering
	\includegraphics[width=0.4\columnwidth]{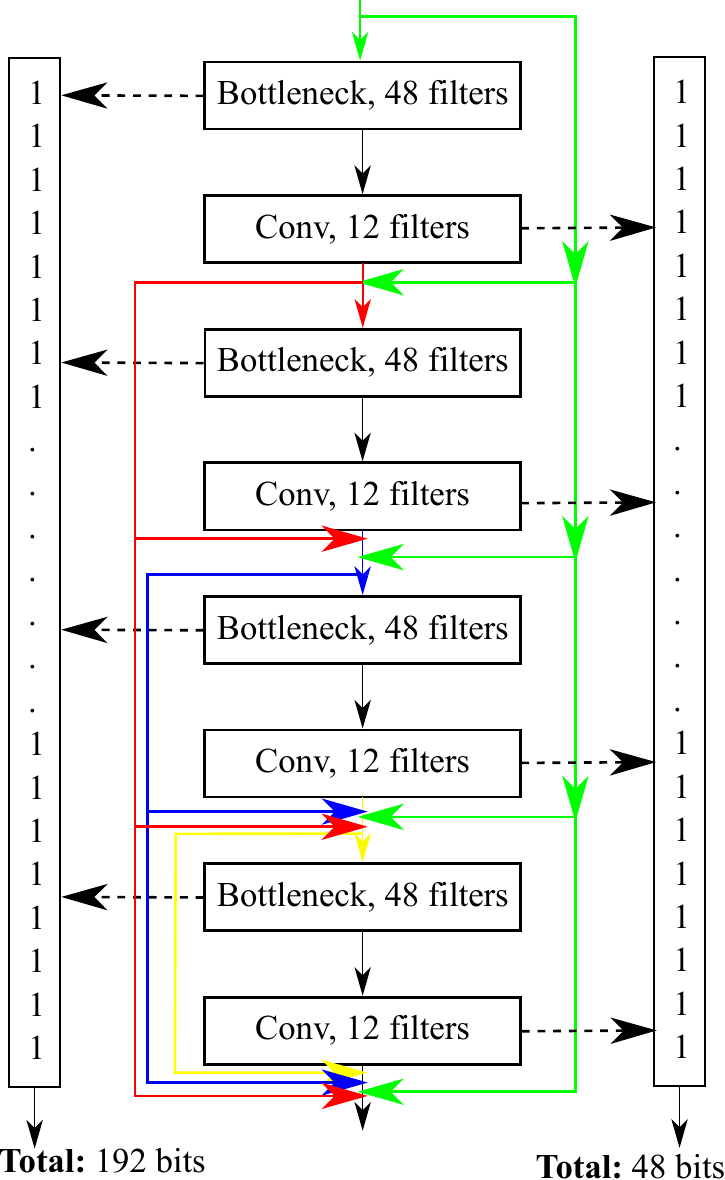}
	\caption{Binary representation of a single dense block used in DenseNet models.}
	\label{fig:densenet-representation}
\end{figure}

However, the representation depends on the type of DCNN model being pruned. When pruning simple CNN models, only a single binary string is used, and each bit will represent a filter of the CNN model, as illustrated in Figs. \ref{fig:CNN-representation1} and \ref{fig:CNN-representation2}. In the case of ResNet models, two binary strings are used to encode the filters: one represents the first convolution layer, while the other represents the second convolution layer of a residual block. Because a tensor entering a residual block needs to be added with a tensor exiting the residual block, see Eq. \ref{eq:resnet}, we must prune the second layer of all residual blocks at the same time in order to maintain the size of all tensors in the model consistent. This process is illustrated in Fig. \ref{fig:resnet-representation} where, in the left side, we have one binary string encoding the first layers of each residual block, and, in the right side, we have one binary string encoding the second layers of multiple residual blocks. Note that the second layers from blocks having the same number of filters, for example 16 filters, are encoded together. For example, in Fig. \ref{fig:resnet-representation}, the layers in green (represented by small bars in the right side of the box) only need 16 bits to be encoded, and, thus, if we eliminate four filters from the first green layer, we will also need to delete four filters from all other green layers. For last, the binary representation used for DenseNet models also uses two binary strings: one to represent the bottleneck layers, and the other to represent the convolution layers, as shown in Fig. \ref{fig:densenet-representation}. Because DenseNets use concatenation instead of addition, it is possible to prune each layer of the model independently, and, therefore, they are easier to represent than ResNets.

\subsection{Population Initialization}
The first step of the proposed \textit{DeepPruningES} is to initialize a population of $3+\lambda_{size}$ individuals. First, all individuals are initialized as an identical copy of the original DCNN model being pruned. Then, their genes are mutated with probability $p_m$ to generate a pruned version of the original model. Finally, the main \textit{for-loop}, shown in Algorithm \ref{alg:DeepPruningES}, line 2, can be executed.

The main idea of initializing the population in this way instead of a purely randomized initialization is that the population will resemble the original DCNN model allowing us to achieve good results even when using smaller populations and fewer generations.

\subsection{Individual Evaluation}

Before we can talk about how the knee and boundaries solutions are selected, it is important to explain how an individual is evaluated in the proposed algorithm. This process is achieved by, first, eliminating the filters indicated in the individual’s gene, and, then, fine tuning the resulting pruned model using Stochastic Gradient Descent (SGD) for $e_{eval}$ epochs and $\alpha_{eval}$ learning rate in a small sample of the dataset that was used to train the original model. This allows the individual to restore some of its performance before the selection operation is performed. More details about how the dataset is sampled and for how long the pruned model is fine-tuned are presented in Subsection \ref{subsection:chosen-dnns}.

The individuals are evaluated in two objectives: the training error in the chosen dataset, and the number of floating-point operations (FLOPs) for a single input. These are conflicting objectives because it is not possible to maintain a small training error with a small number of floating-point operations. We choose to use the number of FLOPs instead of the total number of parameters or the total number of filters because the time to compute the outputs of any given layer is also dependent on the size of its inputs not only on its number of parameters. Thus, the total number of FLOPs of a model is considered a better measurement of its computational complexity.

\subsection{Knee and Boundary Selection}

In many real-world problems, decision makers (DM) have to optimize multiple parameters at the same time to accomplish some desired behavior. These problems are called multi-criteria decision making (MCDM) problems, where there is no single desired or optimal output to a given problem. In many cases, these problems are multi-objective optimization problems where the modification of a single parameter from a candidate solution will change the output of different objective functions. However, not all possible solutions are of interest in MCDM problems. Only the solutions located in the so-called \textit{Pareto optimal front} in the objective space is of interest.  This region is where solutions that are no worse than others in all objective functions are located. The \textit{knee} solution is always found in the \textit{Pareto optimal front}. The ability to locate the \textit{knee} solution is of extreme importance in MCDM problems because this is the solution that can improve the most in any of the objectives with small changes in its parameters \cite{chiu_minimum_2016}.

In the case of DCNNs, the parameters are the number of filters still present in the architecture, and the objective functions that we would like to optimize are the DCNN's computational complexity, and its accuracy in a given dataset. Furthermore, we would like to provide three DCNN models to DMs. So, depending on what type of device the DCNN is being deployed one model could be more suitable than the others.

Commonly, all the solutions located in \textit{Pareto optimal front} need to be found before the \textit{knee} can be identified. Chiu \textit{et al.} in \cite{chiu_minimum_2016} showed that \textit{knee} is the solution in \textit{Pareto optimal front} with the smallest Manhattan distance to the origin point of the objective space. However, finding the whole \textit{Pareto optimal front} is a non-trivial task. Thus, we propose a modification of the Minimum Manhattan Distance algorithm developed by Chiu \textit{et al.} to locate the \textit{knee} solution. Because we are not interested in finding the whole \textit{Pareto optimal front}, we can use the geometric position of all candidate solutions in the objective space to facilitate the search of the \textit{knee} solution. This is done by simply computing the Manhattan distances of all candidate solutions, and finding the one with the smallest distance.

\begin{figure}[!t]
	\centering
	\includegraphics[width=0.7\columnwidth]{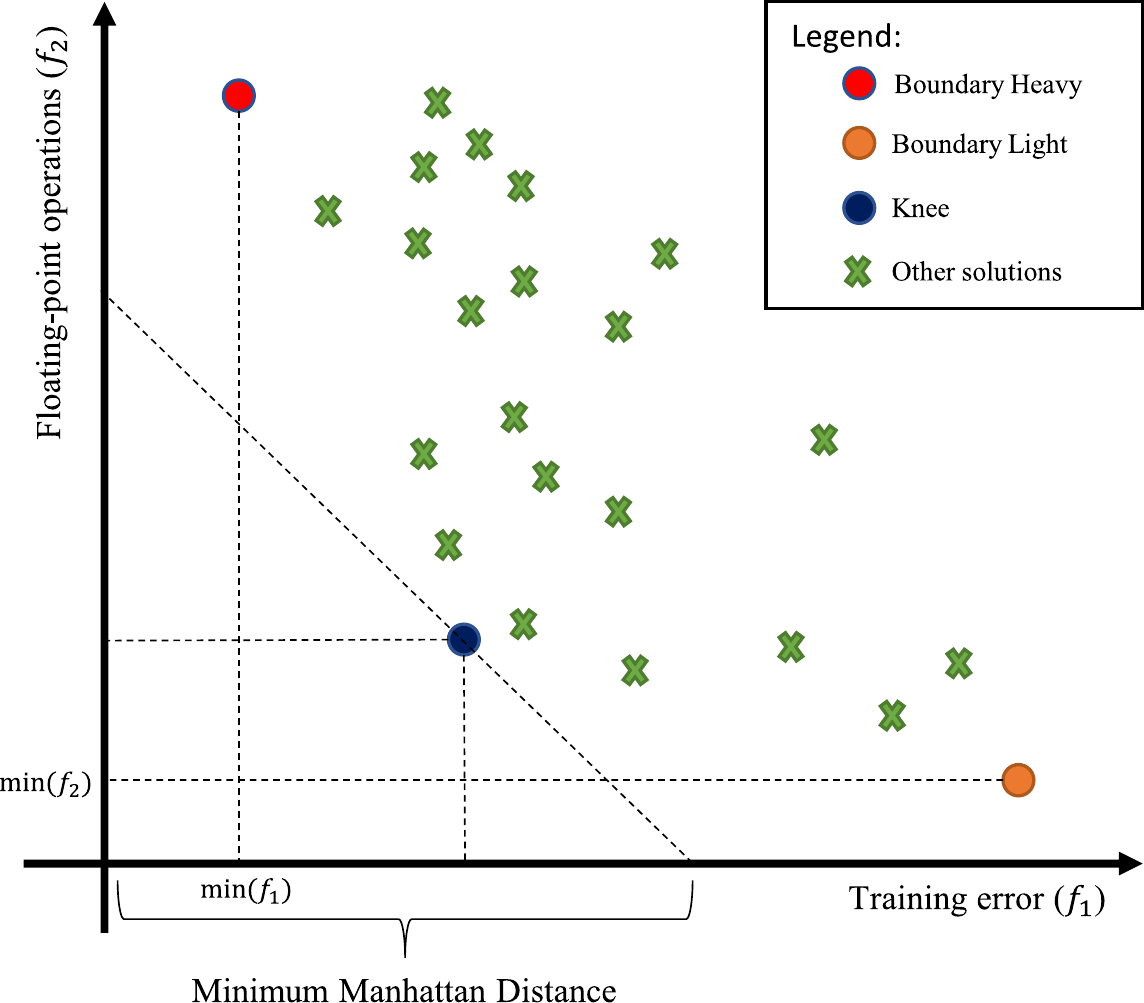}
	\caption{Example of the \textit{knee}, \textit{boundary heavy}, and \textit{boundary light} solutions.}
	\label{fig:MMD}
\end{figure}

Thus, the proposed algorithm to perform the \textit{Knee and Boundary Selection} is shown in Algorihtm \ref{alg:MMD}, where three individuals, called here as \textit{knee}, \textit{boundary heavy}, and \textit{boundary light} solutions, are always selected at every generation. First, we need to evaluate the individuals in the population, $\mathbf{P}$, and to obtain their training error ($f_1$) and number of FLOPs ($f_2$). Second, we need to find the minimum and maximum values of both objectives in the population (Algorithm \ref{alg:MMD}, line 2). Third, the \textit{Boundary Heavy} solution will be the individual with the smallest training error in the population (i.e., the individual with its $f_1$ equal to $min(f_1)$). While the \textit{Boundary Light} solution will be the individual with the smallest number of FLOPs in the population, the individual with its $f_2$ equal to $min(f_2)$. Finally, we compute the Manhattan Distance of all individuals in the population by following the equation in Algorithm \ref{alg:MMD}, line 6, and select the individual with the Minimum Manhattan Distance (MMD) to be our \textit{knee} solution. The selection procedure is illustrated in Fig. \ref{fig:MMD}, and only these three individuals are returned at the end of selection.

\begin{algorithm}[!t]
	\caption{Knee and Boundary Selection}
	\label{alg:MMD}
	\SetKwInOut{Input}{Input}\SetKwInOut{Output}{Output}
	
	\Input{Individuals in the Population ($\mathbf{P}$), original DCNN model ($model$), number of epochs for individual evaluation ($e_{eval}$), learning rate for individual evaluation ($\alpha_{eval}$).}
	\Output{Three individuals: knee solution ($\mu.knee$), boundary heavy solution ($\mu.heavy$) and boundary light solution ($\mu.light$).}
	
	\BlankLine
	\textit{Evaluate Population}($\mathbf{P}$, $model$, $e_{eval}$, $\alpha_{eval}$)\;
	Find $min(f_1), min(f_2), max(f_1),  max(f_2)$\;
	$\mu.heavy \leftarrow P_i$, where $f_1(P_i)=min(f_1)$\;
	$\mu.light \leftarrow P_j$, where $f_2(P_j)=min(f_2)$\;
	
	\For{k = 1 to len($\mathbf{P}$)}
	{
		$ dist(k) = \frac{f_1(P_k) - min(f_1)}{max(f_1) - min(f_1)} + \frac{f_2(P_k) - min(f_2)}{max(f_2) - min(f_2)}$\;
	}
	$\mu.knee \leftarrow P_k$, where $P_k$ has the minimum $dist(k)$\;
	\textbf{return} $\mu.knee$, $\mu.heavy$, $\mu.light$\;
\end{algorithm}

\subsection{Offspring Generation}

After the three individuals from the knee and boundaries are selected, the algorithm will use them to generate new offspring. First, from these three individuals, one is randomly selected to be the parent of a single offspring. Then, each one of its gene elements is randomly flipped with probability equal to $p_m$. This process is repeated until we have an offspring population equal to $\lambda_{size}$. It is also important to point out that we are following the standard ES algorithm where only the knee and boundaries solutions are used to generate a new offspring population. Thus, a newly created offspring is never used to generate another offspring. This method ensures that new offspring are always generated closer to the knee or boundary solutions allowing the algorithm to exploit these regions of interest efficiently. 

\subsection{Fine Tuning}

The last step of the proposed algorithm is to fine-tune the knee and boundaries solutions found in the last generation. They are fine-tuned using Stochastic Gradient Descent (SGD) during $e_{fine}$ epochs and learning rate equal to $\alpha_{fine}$. Different from the Individual Evaluation, the fine-tuning process uses the entire training set in order to improve these three solutions as much as possible. The number of epochs during fine-tuning is also higher than the one used during individual evaluation. At the end of this process, the three solutions are saved on disk for late use.

\section{Experimental Design}

In this section, we discuss the experimental design used to evaluate the proposed \textit{DeepPruningES}. We begin by presenting the DCNN architectures tested with the proposed work. Then, we analyze the algorithm parameters that was used to prune those DCNN architectures. Furthermore, all results presented in this work were obtained using a Laptop PC with a Core i7-8750H CPU, 16 GB of RAM, and a NVIDIA GTX 1060 6GB GPU running PyTorch on Windows 10 Pro 64-bits.

\subsection{Chosen DCNN Architectures for Pruning}
\label{subsection:chosen-dnns}

The proposed \textit{DeepPruningES} is evaluated on state-of-the-art DCNN architectures. As stated in the introduction of this work, the proposed algorithm can prune three types of DCNN architectures: Convolutional Neural Networks (CNN), Residual Neural Networks (ResNet), and Densely Connected Neural Networks (DenseNets). We choose to test the algorithm with DCNNs from these architectures, given a challenging dataset.

\begin{figure}[!t]
	\centering
	\includegraphics[width=0.7\columnwidth]{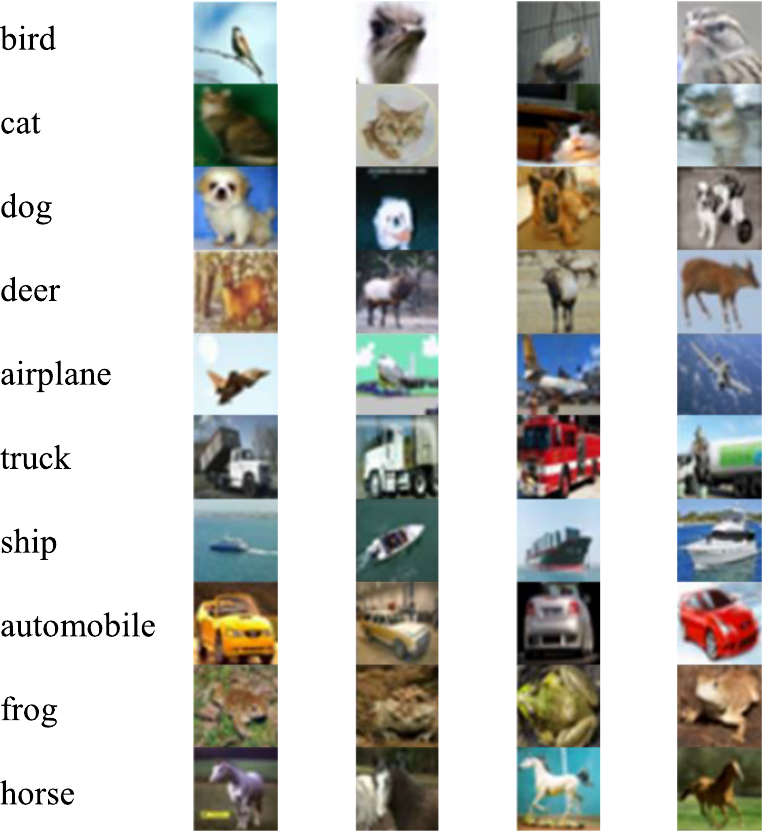}
	\caption{Example of images from each class in the CIFAR10 dataset.}
	\label{fig:CIFAR10}
\end{figure}

The CIFAR10 and CIFAR100 datasets created by Krizhevsky \cite{krizhevsky_learning_2009} are the chosen ones to be used in our experiments because, besides being a challenging dataset, there are many state-of-the-art DCNNs results widely reported to allow us to stay focus on quantifying the quality of the pruning task. The CIFAR10 dataset contains 50,000 training images and 10,000 test images distributed in a total of 10 classes. All images are in colors and have size equal to $32 \times 32$ pixels. Some examples of images found in the CIFAR10 dataset can be seen in Fig. \ref{fig:CIFAR10}. Likewise, the CIFAR100 dataset contains the same amount of images with the same resolution as the CIFAR10, but these images are distributed in a total of 100 classes instead of only 10. To speed up and reduce the cost of the pruning process, only 1,000 training images, with an equal number of images per class, from the CIFAR10 and CIFAR100 datasets are used to perform an \textit{Individual Evaluation}. At the end of the proposed algorithm, when fine-tuning the three selected individuals, the whole 50,000 training images from the these datasets are used to allow the candidate solutions to recover from any loss of performance due to the pruning process.

We choose to evaluate the proposed algorithm with the following networks: \textit{VGG16}, \textit{VGG19}, \textit{ResNet56}, \textit{ResNet110}, \textit{DenseNet50}, and \textit{DenseNet100}. The \textit{VGG16} and \textit{VGG19} \cite{simonyan_very_2014} are two examples of CNNs created in 2014 which achieved good results in challenging image classifications datasets, such as CIFAR10 and ImageNet. As the name suggests, \textit{VGG16} has a total of 16 layers while \textit{VGG19} has a total of 19 layers. The \textit{VGG16} and \textit{VGG19} have computational complexities of $3.15 \times 10^8$ and $4.01 \times 10^8$ FLOPs, respectively. Their original and fully trained architectures can achieve a test error of $6.06\%$ and $6.18\%$ on the CIFAR10 dataset, respectively, and $32.33\%$ and $32.57\%$ test error on the CIFAR100 dataset, respectively. The \textit{ResNet56} and \textit{ResNet110} \cite{he_deep_2016} are two example of ResNets, and they are the ones first proposed by the authors in \cite{he_deep_2016} to be used with the CIFAR10 dataset. Their computational complexities are of $1.27 \times 10^8$ and $2.57 \times 10^8$ FLOPs, respectively. When fully trained, \textit{ResNet56} and \textit{ResNet110} can achieve a test error of $6.63\%$ and $6.2\%$ on CIFAR10, respectively, and $41.94\%$ and $50.47\%$ on the CIFAR100 dataset, respectively. The \textit{ResNet56} contains a total of 56 layers, while the \textit{ResNet110} contains a total of 110 layers. For last, the \textit{DenseNet50} and \textit{DenseNet100} \cite{huang_densely_2017} are also some of the proposed architectures based on the DenseNet’s authors to classify the CIFAR10 dataset. They have 50 and 100 layers, respectively, computational complexity of $0.93 \times 10^8$ and $3.05 \times 10^8$ FLOPs, respectively, and test error of $6.92\%$ and $5.66\%$ on the CIFAR10 dataset, respectively, and $41.25\%$ and $33.06\%$ on the CIFAR100 dataset, respectively. Table \ref{table:CIFAR10} sums up the details about the DCNN architectures used to evaluate the proposed algorithm.

\begin{table}[!t]
	\caption{Overview of the DCNN architectures used to evaluate the proposed \textit{DeepPruningES}.}
	\label{table:CIFAR10}
	\centering
	\begin{adjustbox}{width=\columnwidth,center}
	\begin{tabular}{c|c|c|c|c}
		\hline
		\textbf{DCNN} & \textbf{\# layers} &\textbf{ \# FLOPs} & \textbf{Test error on CIFAR10} & \textbf{Test error on CIFAR100}\\
		\hline
		\hline
		VGG16 & 16 & $3.15 \times 10^8$	& $6.06\%$ & $32.33\%$\\ \hline
		VGG19 & 19 & $4.01 \times 10^8$ & $6.18\%$ & $32.57\%$\\ \hline
		ResNet56 & 56 & $1.27 \times 10^8$ & $6.63\%$ & $41.94\%$\\ \hline
		ResNet110 & 110 & $2.57 \times 10^8$ & $6.2\%$ & $50.47\%$\\ \hline
		DenseNet50 & 50 & $0.93 \times 10^8$ & $6.92\%$ & $41.25\%$\\ \hline
		DenseNet100 & 100 & $3.05 \times 10^8$ & $5.66\%$ & $33.06\%$\\ \hline
	\end{tabular}
	\end{adjustbox}
\end{table}

\subsection{Algorithm Parameters}

The parameters’ values used to evaluate the proposed \textit{DeepPruningES} algorithm are shown in Table \ref{table:parameters}. Throughout this study, all results shown related to the proposed algorithm are obtained by using the parameters from Table \ref{table:parameters}. If some results are obtained by using a different set of parameters, the difference is highlighted in the discussions. The next paragraphs will discuss the influence of each parameter in the final output of the \textit{DeepPruningES}.

The size of the offspring ($\lambda_{size}$) defines how many offspring is generated at each generation. More offspring helps the algorithm in covering more of the objective space and increases the probability of finding good solutions. Likewise, a large number of generations ($gen$) also increases the probability of finding good solutions. However, when the number of offspring and/or generation increases, the computational cost to run the algorithm also increases. In our tests, $20$ offspring and $10$ generations are considered enough to achieve good results given affordable computational complexity.

The mutation probability ($p_m$) is the probability of a single bit of an individual’s gene being inverted. A very small mutation probability would take too long to produce significant changes in the population, which would require more individuals and more generations to reach a satisfactory result; while a very large mutation probability will produce too many changes in the population resulting in no improvement at all.. In our experiments, a probability of $0.1$ is regarded as the best value heuristically that makes the individuals change in a way that improves the overall quality of the population.

The number of epochs and learning rate used during an individual evaluation ($e_{eval}$, $\alpha_{eval}$) will control how long an individual will be fine-tuned before its training error and number of FLOPs is evaluated. The values for these two parameters depend on the computational power available to the user. Ideally, if an individual can be fine-tuned much longer, this will give us a better sense of its overall fitness. We chose to use $5$ epochs, which is compatible with other EC based algorithms used for DCNNs architecture generation and pruning \cite{luo_thinet:_2017, luo_entropy-based_2017, fernandes_jr._particle_2019}, and a learning rate of $0.1$. In our case, we are using a large learning rate to improve the accuracy of the solutions faster.

For last, the number of epochs and learning rate to fine-tune the final solutions ($e_{fine}$, $\alpha_{fine}$) will improve the training error of the three final solutions. We chose to fine-tune them for $50$ epochs with a learning rate of $0.01$, which is also compatible with others' work \cite{li_pruning_2017, ding_auto-balanced_2018}. In this case, we use a smaller learning rate than that during an individual evaluation to avoid training instabilities due to excessive removed filters. It is possible to improve the solutions found by the algorithm even further if they are fine-tuned for longer than $50$ epochs. However, our limited computational power did not allow us to experiment with very long fine-tuning sessions.

As stated before, the proposed \textit{DeepPruningES} uses the ES \textit{plus} version for performing DCNN filter pruning, which is an elitist version of ES where the best individuals are never lost. In Section V, an experiment using the ES \textit{comma} version is presented, and it does not yield good results because the algorithm eliminates too many filters even from the boundary heavy and knee solutions.

\begin{table}[!t]
	\caption{Parameters used to evaluate the proposed \textit{DeepPruningES}.}
	\label{table:parameters}
	\centering
	\begin{tabular}{c|c}
		\hline
		\textbf{Parameter} & \textbf{Value} \\
		\hline
		\hline
		Offspring size ($\lambda_{size}$) & $20$\\ \hline
		Maximum number of generations ($gen$)	& $10$ \\ \hline
		Mutation probability ($p_m$) & $0.1$ \\ \hline
		Number of epochs for individual evaluation ($e_{eval}$) & $5$ \\ \hline
		Learning rate for individual evaluation ($\alpha_{eval}$) & $0.1$ \\ \hline
		Number of epochs for fine-tuning ($e_{fine}$) & $50$ \\ \hline
		Learning rate for fine-tuning ($\alpha_{fine}$) & $0.01$ \\ \hline
	\end{tabular}
\end{table}

\section{Experimental Results and Discussion}

In this section, our pruning results are first presented and compared with peer competitors. Then, a detailed discussion about the obtained results follows.

\subsection{Experimental Results}
To evaluate the proposed \textit{DeepPruningES}, the DCNN architectures shown in Table \ref{table:CIFAR10} were pruned $10$ times. The results using the CIFAR10 dataset are shown in Table \ref{table:results_CIFAR10}, where, for each DCNN architecture, we show the best and mean test error and number of FLOPs for the knee, boundary heavy, and boundary light solutions after fine-tuning for 50 epochs. The results using the CIFAR100 dataset are shown in Table \ref{table:results_CIFAR100}, where the three selected solution are fine-tuned for 100 epochs. In addition, in Table \ref{table:competitors}, we show the results reported by the peer competitors who used the same CIFAR10 dataset. Other works found in the literature are not shown here because they used the ImageNet dataset, which is too big for us to run in our equipment and does not produce comparable results with other datasets.

\begin{table*}[!t]
	\caption{Pruning results obtained with the proposed \textit{DeepPruningES} on the CIFAR10 dataset.}
	\label{table:results_CIFAR10}
	\centering
	\begin{adjustbox}{max width=\textwidth,center}
	\begin{tabular}{c|c|c|c|c|c|c|c}
		\hline
		\textbf{DCNN Model} & \multicolumn{7}{c}{\textbf{DeepPruningES}} \\
		\hline
		\hline
		\multirow{4}{*}{\textbf{VGG16}} & \textbf{Solution} & \textbf{Test error (best)} & \textbf{Test error (mean)} & \textbf{\# FLOPs (best)} & \textbf{\# FLOPs (mean)} & \textbf{\% Pruned (best)} & \textbf{\% Pruned (mean)} \\

		~ & \textbf{Knee} & $9.04\%$ & $9.58\%$ & $1.09 \times 10^8$ & 	$1.22 \times 10^8$ & $65.49\%$ & $61.58\%$ \\
		
		~ & \textbf{Boundary Heavy} & $8.21\%$ & $8.6\%$ & $2.15 \times 10^8$ & $2.49 \times 10^8$ & $32.01\%$ & $20.88\%$\\
		~ & \textbf{Boundary Light} & $10.51\%$ & $11.41\%$ & $0.88 \times 10^8$ & $0.9 \times 10^8$ & $72.17\%$ & $71.36\%$\\
		\hline
		
		\multirow{4}{*}{\textbf{VGG19}} & \textbf{Solution} & \textbf{Test error (best)} & \textbf{Test error (mean)} & \textbf{\# FLOPs (best)} & \textbf{\# FLOPs (mean)} & \textbf{\% Pruned (best)} & \textbf{\% Pruned (mean)} \\
		
		~ & \textbf{Knee} & $9.04\%$ & $9.87\%$ & $1.53 \times 10^8$ & $1.72 \times 10^8$ & $61.86\%$ & $57.15\%$\\
		~ & \textbf{Boundary Heavy} & $8.21\%$ & $8.77\%$ & $2.7 \times 10^8$ & $3.12\times 10^8$ & $32.56\%$ & $22.28\%$\\
		~ & \textbf{Boundary Light} & $10.53\%$ & $12.03\%$ & $1.13 \times 10^8$ & $1.18\times 10^8$ & $71.74\%$ & $70.69\%$\\
		\hline
		
		\multirow{4}{*}{\textbf{ResNet56}} & \textbf{Solution} & \textbf{Test error (best)} & \textbf{Test error (mean)} & \textbf{\# FLOPs (best)} & \textbf{\# FLOPs (mean)} & \textbf{\% Pruned (best)} & \textbf{\% Pruned (mean)} \\
		
		~ & \textbf{Knee} & $9.28\%$ & $9.98\%$ & $0.432 \times 10^8$ & $0.523 \times 10^8$ & $66.23\%$ & $59.15\%$ \\
		~ & \textbf{Boundary Heavy} & $8.11\%$ & $8.77\%$ & $1.01 \times 10^8$ & $1.08 \times 10^8$ & $21.31\%$ & $15.23\%$ \\
		~ & \textbf{Boundary Light} & $11.42\%$ & $13.36\%$ & $0.244 \times 10^8$ & $0.286 \times 10^8$ & $80.89\%$ & $77.67\%$ \\
		\hline
		
		\multirow{4}{*}{\textbf{ResNet110}} & \textbf{Solution} & \textbf{Test error (best)} & \textbf{Test error (mean)} & \textbf{\# FLOPs (best)} & \textbf{\# FLOPs (mean)} & \textbf{\% Pruned (best)} & \textbf{\% Pruned (mean)} \\
		
		~ & \textbf{Knee} & $8.66\%$ & $9.42\%$ & $0.905 \times 10^8$ & $1.03 \times 10^8$ & $64.84\%$ & $59.89\%$ \\
		~ & \textbf{Boundary Heavy} & $7.43\%$ & $7.93\%$ & $2.14 \times 10^8$ & $2.21\times 10^8$ & $16.72\%$ & $14.14\%$ \\
		~ & \textbf{Boundary Light} & $10.27\%$ & $12.9\%$ & $0.43 \times 10^8$ & $0.56 \times 10^8$ & $83.29\%$ & $77.86\%$ \\
		\hline
		
		\multirow{4}{*}{\textbf{DenseNet50}} & \textbf{Solution} & \textbf{Test error (best)} & \textbf{Test error (mean)} & \textbf{\# FLOPs (best)} & \textbf{\# FLOPs (mean)} & \textbf{\% Pruned (best)} & \textbf{\% Pruned (mean)} \\
		
		~ & \textbf{Knee} & $9.8\%$ & $10.43\%$ & $0.41 \times 10^8$ & $0.466 \times 10^8$ & $56.05\%$ & $50.15\%$ \\
		~ & \textbf{Boundary Heavy} & $8.91\%$ & $9.26\%$ & $0.756 \times 10^8$ & $0.779 \times 10^8$ & $19.16\%$ & $16.59\%$ \\
		~ & \textbf{Boundary Light} & $13.04\%$ & $14.8\%$ & $0.229 \times 10^8$ & $0.244 \times 10^8$ & $75.53\%$ & $73.91\%$ \\
		\hline
		
		\multirow{4}{*}{\textbf{DenseNet100}} & \textbf{Solution} & \textbf{Test error (best)} & \textbf{Test error (mean)} & \textbf{\# FLOPs (best)} & \textbf{\# FLOPs (mean)} & \textbf{\% Pruned (best)} & \textbf{\% Pruned (mean)} \\
		
		~ & \textbf{Knee} & $9.04\%$ & $9.37\%$ & $1.11 \times 10^8$ & $1.21 \times 10^8$ & $63.64\%$ & $60.31\%$ \\
		~ & \textbf{Boundary Heavy} & $8.34\%$ & $8.39\%$ & $2.46 \times 10^8$ & $2.49 \times 10^8$ & $19.33\%$ & $18.24\%$ \\
		~ & \textbf{Boundary Light} & $10.47\%$ & $11.90\%$ & $0.82 \times 10^8$ & $0.879 \times 10^8$ & $73.09\%$ & $71.16\%$ \\
		\hline
	\end{tabular}
	\end{adjustbox}
\end{table*}

\begin{table*}[!t]
	\caption{Pruning results obtained with the proposed \textit{DeepPruningES} on the CIFAR100 dataset.}
	\label{table:results_CIFAR100}
	\centering
	\begin{adjustbox}{max width=\textwidth,center}
		\begin{tabular}{c|c|c|c|c|c|c|c}
			\hline
			\textbf{DCNN Model} & \multicolumn{7}{c}{\textbf{DeepPruningES}} \\
			\hline
			\hline
			\multirow{4}{*}{\textbf{VGG16}} & \textbf{Solution} & \textbf{Test error (best)} & \textbf{Test error (mean)} & \textbf{\# FLOPs (best)} & \textbf{\# FLOPs (mean)} & \textbf{\% Pruned (best)} & \textbf{\% Pruned (mean)} \\
			
			~ & \textbf{Knee} & 34.27\% & 34.41\% & $1.289 \times 10^8$ & $1.411 \times 10^8$ & 59.19\% & 55.37\%\\
			~ & \textbf{Boundary Heavy} & 32.94\% & 33.08\% & $2.531 \times 10^8$ & $2.577 \times 10^8$ & 19.93\% & 18.47\%\\
			~ & \textbf{Boundary Light} & 35.68\% & 36.36\% & $6.786 \times 10^7$ & $7.015 \times 10^7$ & 78.53\% & 77.81\%\\
			\hline
			
			\multirow{4}{*}{\textbf{VGG19}} & \textbf{Solution} & \textbf{Test error (best)} & \textbf{Test error (mean)} & \textbf{\# FLOPs (best)} & \textbf{\# FLOPs (mean)} & \textbf{\% Pruned (best)} & \textbf{\% Pruned (mean)} \\
			
			~ & \textbf{Knee} & 34.6\% & 35.09\% & $1.652 \times 10^8$ & $1.882 \times 10^8$ & 58.82\% & 53.08\%\\
			~ & \textbf{Boundary Heavy} & 33.11\% & 33.41\% & $3.246 \times 10^8$ & $3.277 \times 10^8$ & 19.09\% & 18.31\%\\
			~ & \textbf{Boundary Light} & 37.05\% & 37.41\% & $8.640 \times 10^7$ & $8.970 \times 10^7$ & 78.46\% & 77.64\%\\
			\hline
			
			\multirow{4}{*}{\textbf{ResNet56}} & \textbf{Solution} & \textbf{Test error (best)} & \textbf{Test error (mean)} & \textbf{\# FLOPs (best)} & \textbf{\# FLOPs (mean)} & \textbf{\% Pruned (best)} & \textbf{\% Pruned (mean)} \\
			
			~ & \textbf{Knee} & 53.98\% & 55.57\% & $1.484 \times 10^7$ & $1.681 \times 10^7$ & 88.40\% & 86.86\%\\
			~ & \textbf{Boundary Heavy} & 42.19\% & 42.49\% & $1.072 \times 10^8$ & $1.096 \times 10^8$ & 16.19\% & 14.26\% \\
			~ & \textbf{Boundary Light} & 53.89\% & 56.70\% & $1.421 \times 10^7$ & $1.587 \times 10^7$ & 88.89\% & 87.61\%\\
			\hline
			
			\multirow{4}{*}{\textbf{ResNet110}} & \textbf{Solution} & \textbf{Test error (best)} & \textbf{Test error (mean)} & \textbf{\# FLOPs (best)} & \textbf{\# FLOPs (mean)} & \textbf{\% Pruned (best)} & \textbf{\% Pruned (mean)} \\
			
			~ & \textbf{Knee} & 59.06\% &  64.09\% & $3.068 \times 10^7$ & $3.633 \times 10^7$ & 88.08\% & 85.89\%\\
			~ & \textbf{Boundary Heavy} & 50.97\% & 51.49\% & $2.117 \times 10^8$ & $2.175 \times 10^8$ & 17.73\% & 15.63\%\\
			~ & \textbf{Boundary Light} & 62.07\% & 64.93\% & $3.069 \times 10^7$ & $3.365 \times 10^7$ & 88.07\% & 86.93\%\\
			\hline
			
			\multirow{4}{*}{\textbf{DenseNet50}} & \textbf{Solution} & \textbf{Test error (best)} & \textbf{Test error (mean)} & \textbf{\# FLOPs (best)} & \textbf{\# FLOPs (mean)} & \textbf{\% Pruned (best)} & \textbf{\% Pruned (mean)} \\
			
			~ & \textbf{Knee} & 42.09\% & 62.01\% & $1.606 \times 10^7$ & $2.643 \times 10^7$ & 82.82\% & 71.73\%\\
			~ & \textbf{Boundary Heavy} & 41.59\% & 42.02\% & $7.569 \times 10^7$ &  $7.815 \times 10^7$ & 19.05\% & 16.42\%\\
			~ & \textbf{Boundary Light} & 63.44\% & 65.06\% & $1.606 \times 10^7$ & $1.705 \times 10^7$ & 82.82\% & 81.76\%\\
			\hline
			
			\multirow{4}{*}{\textbf{DenseNet100}} & \textbf{Solution} & \textbf{Test error (best)} & \textbf{Test error (mean)} & \textbf{\# FLOPs (best)} & \textbf{\# FLOPs (mean)} & \textbf{\% Pruned (best)} & \textbf{\% Pruned (mean)} \\
			
			~ & \textbf{Knee} & 50.46\% & 49.13\% & $6.351 \times 10^7$ & $6.503 \times 10^7$ & 79.18\% & 78.69\%\\
			~ & \textbf{Boundary Heavy} & 35.344\% & 34.41\% & $2.469 \times 10^8$ & $2.50 \times 10^8$ & 19.05\% & 18.04\%\\
			~ & \textbf{Boundary Light} & 50.24\% & 49.24\% & $6.185 \times 10^7$ & $6.316 \times 10^7$ & 79.73\% & 79.29\%\\
			\hline
		\end{tabular}
	\end{adjustbox}
\end{table*}

On the CIFAR10 dataset, for the \textit{VGG16} and \textit{VGG19}, the proposed pruning algorithm obtain similar results. The boundary heavy solutions have an average of $20\%$ reduction in the number of FLOPs, the boundary light solutions an average of $70\%$ reduction, and the knee solutions an average of $57\%$ reduction. The average test errors of the boundary heavy, boundary light, and knee solutions are around $8.6\% \sim 8.77\%$, $11.51\% \sim 12.01\%$, and $9.58\% \sim 9.87\%$, respectively. These are the expected results where the \textit{boundary heavy} solutions were the least pruned, and also the ones with the highest accuracy. On the other hand, the \textit{boundary light} solutions were the most pruned, and the ones with the lowest accuracy. For last, the \textit{knee} solutions obtained pruned results and accuracy in between.

On the CIFAR10 dataset, for the \textit{ResNet56} and \textit{ResNet110}, the boundary heavy solutions see an average of $15\%$ reduction in the number of FLOPs, the boundary light solutions an average of $77\%$ reduction in the number of FLOPs, and the knee solutions an average of $59\%$ reduction in the number of FLOPs. After fine-tuning, the boundary heavy, boundary light, and knee solutions obtain test errors on average of $7.93\% \sim 8.77\%$, $12.9\% \sim 13.46\%$, and $9.42\% \sim 9.98\%$, respectively. Even though ResNets are smaller than conventional CNNs, such as VGG16 and VCGG19, by nature, these results show that they can still be pruned without losing too much accuracy.

On the CIFAR10 dataset, for the DenseNet50 and DenseNet100, the average reduction in the number of FLOPs for the boundary heavy, boundary light and knee solutions are around $16\%$, $73\%$, and $50\%$, respectively. On the other hand, the average test errors for the boundary heavy, boundary light and knee solutions are $8.39\% \sim 9.26\%$, $11.9\% \sim 14.8\%$, and $9.37\% \sim 10.43\%$, respectively. These results are similar to the ones obtained with the pruning of ResNets which are impressive considering that DenseNets are even smaller than ResNets.

\begin{table}[!t]
	\caption{Pruning results from peer competitors using the CIFAR10 dataset.}
	\label{table:competitors}
	\centering
	\begin{tabular}{c|c|c|c}
		\hline
		\textbf{Approach} & \textbf{DCNN Model} & \textbf{\% FLOPs Pruned} & \textbf{Test error} \\
		\hline
		\hline
		\multirow{3}{*}{Li \textit{et al.} \cite{li_pruning_2017}} & VGG16 & $34.2\%$ & $6.60\%$ \\
		~ & ResNet56 & $27.6\%$ & $6.94\%$ \\
		~ & ResNet110 & $38.6\%$ & $6.70\%$\\
		\hline
		\multirow{2}{*}{Ding \textit{et al.} \cite{ding_auto-balanced_2018}} & VGG16 & $81.39\%$ & $7.56\%$ \\
		~ &  ResNet56 & $66.88\%$ & $9.43\%$ \\
		\hline
	\end{tabular}
\end{table}

\begin{table}[!t]
		\caption{Efficient mobile deep convolutional neural networks trained from scratch results.}
		\label{table:results_mobilenets}
		\centering
		\begin{adjustbox}{max width=\columnwidth,center}
			\begin{tabular}{c|c}
				\hline
				
				~ & \textbf{Test error on CIFAR100} \\
				\hline
				\textbf{MobileNet v2} & 30.91\% \\
				\textbf{SqueezeNet} & 44.53\% \\
				\textbf{EfficientNet-B0} & 48.02\% \\
				\hline
			\end{tabular}
		\end{adjustbox}
\end{table}

\subsection{Results Discussion}

As shown by the results in Tables \ref{table:results_CIFAR10} and \ref{table:results_CIFAR100}, the proposed pruning algorithm can reduce the number of FLOPs in DCNN architectures significantly while preserving good test errors. As shown in Table \ref{table:competitors}, when comparing our results with the ones obtained by Li \textit{et al.} in \cite{li_pruning_2017}, the proposed \textit{DeepPruningES} obtains higher pruning ratios with slight worse test errors on CIFAR10. Ding et al. in \cite{ding_auto-balanced_2018} were able to obtained better pruning ratios and test errors than the ones from the proposed algorithm. However, in their algorithm, the DCNNs needs to be trained with strong and specialized regularization. The proposed \textit{DeepPruningES} uses only the information about the DCNN classification performance and computational complexity to prune it. Thus, without using any regularization or extra knowledge about the DCNN being pruned, \textit{DeepPruningES} is still capable of achieving quality results comparable with the ones from Ding et al. in \cite{ding_auto-balanced_2018}. Moreover, Table \ref{table:results_mobilenets} shows the test errors on the CIFAR100 dataset of multiple efficient mobile DCNNs trained for 150 epochs, and the results show that the pruned version of full-fledged DCNN architecture can achieve competitive classification performance compared with their mobile counterparts. Beyond the qualitative results, the proposed \textit{DeepPruningES} algorithm is capable of finding pruned DCNN models in a couple of hours in an Nvidia GTX 1060 6GB GPU. However, its bottleneck comes once the pruned models are fine-tuned (Algorithm \ref{alg:DeepPruningES}, line 7) using the whole dataset, which, depending on the DCNN model being pruned, could take a couple of days.

It is also important to point out that the primary goal of the proposed algorithm is to find a set of three pruned DCNN models. Thus, a Decision-Maker (DM) has, in fact, four possible options to choose from: the original DCNN model, the \textit{boundary heavy} solution, the \textit{boundary light} solution, or the \textit{knee} solution. Therefore, although the \textit{boundary heavy} solution can look unappealing at first glance, it is still an essential candidate solution that a DM can use depending on his or her available hardware.

\begin{figure}[!t]
	\centering
	\includegraphics[width=0.7\columnwidth]{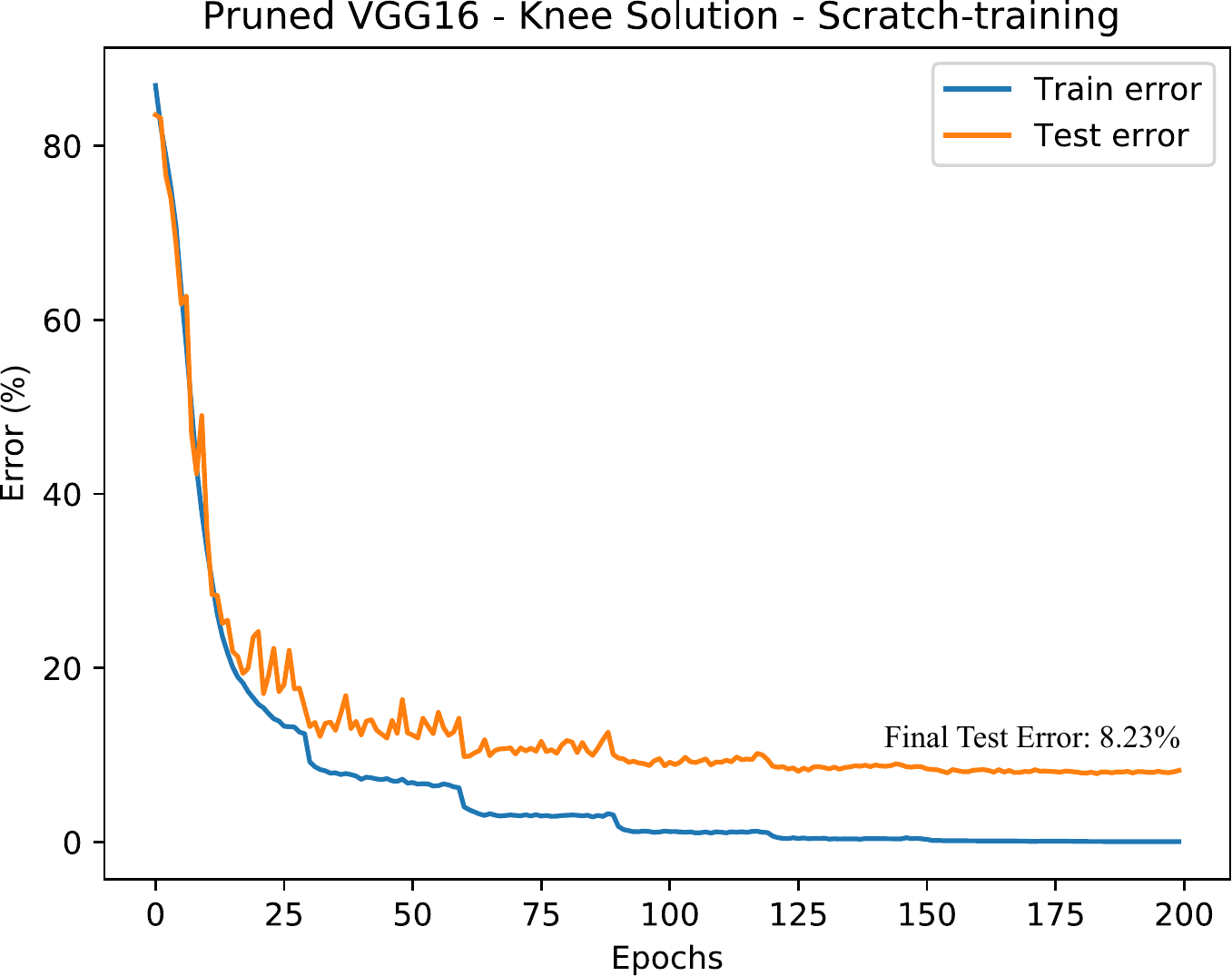}
	\caption{Knee solution from the pruned VGG16 network trained from scratch for 200 more epochs.}
	\label{fig:knee-scratch-training}
\end{figure}

The test errors showed in Tables \ref{table:results_CIFAR10} and \ref{table:results_CIFAR100} can still be further improved given more computational resources are available. To demonstrate this, we choose one pruned solution from VGG16 to be retrained from scratch on the CIFAR10 dataset, which means that all weights are reinitialized to random values and the network is trained from there. More specifically, a knee solution with $58.64\%$ pruning ratio and test error of $9.94\%$ is chosen to be retrained from scratch for 200 more epochs. The training curves for this knee solution is shown in Fig. \ref{fig:knee-scratch-training}. Its final test error is equal to $8.23\%$ which is considerable better than the test error obtained after $50$ fine-tuning epochs.

Our results are more comprehensive than the ones from peer competitors because six different DCNN models with three different architectures and number of layers are evaluated. For example, to the best of our knowledge, no work in the literature has attempted to prune DenseNet architectures. Although DenseNets already have smaller computational complexity with better classification performance than standard CNNs, our results showed that they can still be pruned and still maintain competitive classification performance.

\begin{figure*}[!t]
	\centering
	\includegraphics[width=0.95\textwidth]{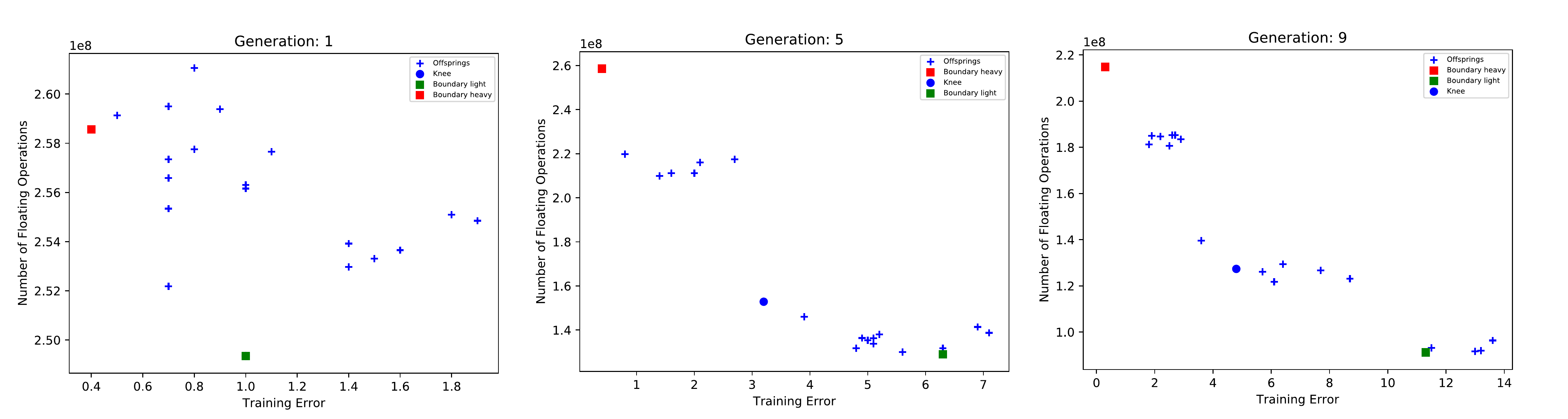}
	\caption{Evolution of the population when pruning VGG16 using the \textit{plus} version of the proposed DeepPruningES.}
	\label{fig:population-plus}
\end{figure*}

\begin{figure*}[!t]
	\centering
	\includegraphics[width=0.95\textwidth]{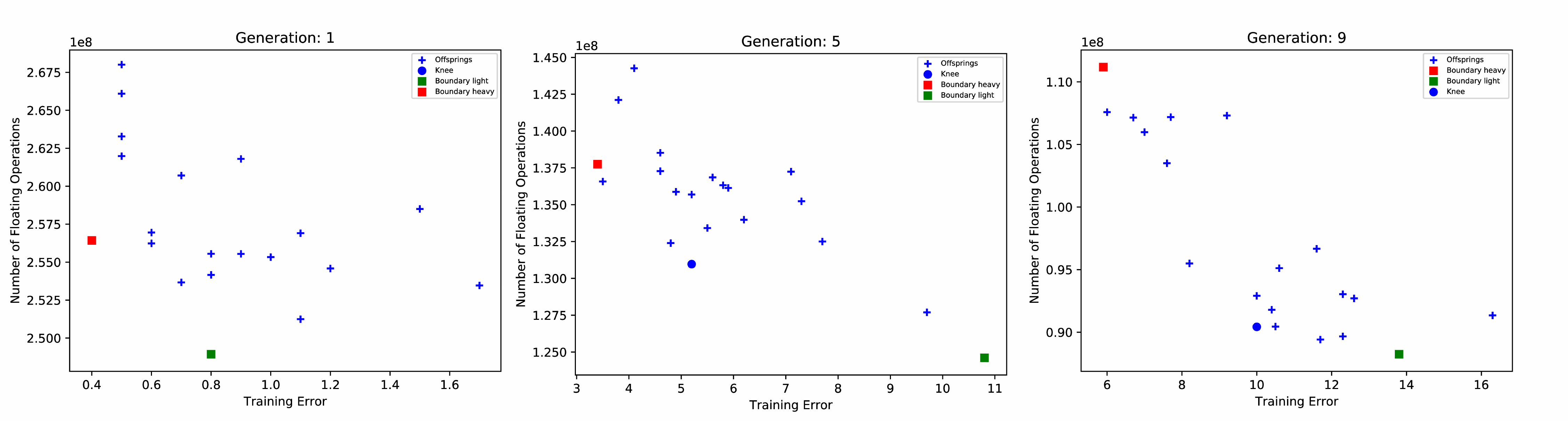}
	\caption{Evolution of the population when pruning VGG16 using the \textit{comma} version of the proposed DeepPruningES.}
	\label{fig:population-comma}
\end{figure*}

We also compared the proposed \textit{DeepPruningES}, which is based on the elitist ES \textit{plus} version, with a non-elitist \textit{comma} version. The evolution of the population for 10 generations for the \textit{plus} version is shown in Fig. \ref{fig:population-plus} and for the \textit{comma} version is shown in Fig. \ref{fig:population-comma}. In the proposed \textit{plus} version, it is easier to keep the individual with the best training error throughout the generations than in the \textit{comma} version. Moreover, the final solutions found when using the \textit{comma} version are very close to each other in terms of error and number of FLOPs which defeats our objective of getting multiple solutions with different trade-offs. Thus, if one is going to use the proposed \textit{DeepPruningES} for pruning DCNN architectures, the plus version of the algorithm is the suggested one to use.

For last, we would like to discuss why an Evolution Strategy (ES) framework was chosen instead of a standard Genetic Algorithm (GA) one. We wanted to give a set of candidate solutions to Decision-Makers (DMs) to choose from instead of a single one. Thus, by maintaining only the three best individuals in the population, the proposed selection operator fits in an ES framework without the developing a fully fledged Multi-Objective GA. Furthermore, although ES was originally designed to deal with real-valued vectors, it is perfectly suitable for use with binary vectors \cite{beyer_evolution_2002}.

\section{Conclusions and Future Work}
In this work, we propose an algorithm for pruning Deep Convolutional Neural Networks (DCNN) architectures using Evolution Strategies, called here as \textit{DeepPruningES}. The pruning is performed by eliminating convolution filters from multiples layers throughout the network architecture. While other works use the current layer’s statistics to decide which filter to eliminate, the proposed algorithm eliminates filters by exploiting the objective space of each solution in the population. Moreover, to the best of our knowledge, the current work is one of the firsts to frame the pruning problem as a multi-objective optimization problem. In this sense, the proposed algorithm can find three different solutions with different trade-offs between training/test error and computational complexity without relying on any prior knowledge from the architecture being pruned. This set of solutions eliminates the need of manually setting a trade-off parameter as seen others' pruning algorithms.

The proposed \textit{DeepPruningES} can achieve competitive results with a population of only 20 individuals and 10 generations when compared with state-of-the-art peer competitors work. The algorithm is also one of the firsts to pruning multiples DCNN architectures, such as CNNs, ResNets, and DenseNets architectures. Our results are limited only by our available computational power. With more computational resources available, the proposed algorithm could achieve even better results.

For future works, the proposed algorithm will be extended to prune more DCNN architectures, such as the Inception architecture, and we will develop an algorithm for searching neural network architectures automatically and integrate it with the proposed pruning algorithm. This will create an algorithm capable of finding new DCNN architectures with good performance while maintaining a small computational complexity.

\section{Acknowledgments}
The authors would like to thank the Brazilian National Council for Scientific and Technological Development (CNPq), grant 203076/2015-0, for the full financial support used to complete this work. Francisco E. Fernandes Jr. also would like to thank the São Paulo Research Foundation (FAPESP) for being awarded a Post-Doctoral Fellowship, grant 2020/05426-0, to support his future academic endeavors.

\bibliography{references}
\end{document}